%% file: acl_latex.tex
\documentclass[11pt]{article}

\usepackage[preprint]{acl}

\usepackage{times}
\usepackage{latexsym}
\newcommand\blfootnote[1]{%
  \begingroup
  \renewcommand\thefootnote{}\footnote{#1}%
  \addtocounter{footnote}{-1}%
  \endgroup
}
\usepackage[T1]{fontenc}

\usepackage[utf8]{inputenc}

\usepackage{microtype}

\usepackage{inconsolata}

\usepackage{graphicx}
\usepackage{amsmath}
\usepackage{booktabs}
\usepackage{tabularx}
\usepackage[most]{tcolorbox}
\usepackage[table]{xcolor}
\usepackage{multirow}
\usepackage{pifont}
\definecolor{yellowhighlight}{RGB}{255,255,204}
\definecolor{bluehighlight}{RGB}{221,235,247}
\definecolor{lightgray}{gray}{0.9}
\definecolor{graytext}{gray}{0.5}
\definecolor{greenhighlight}{RGB}{210,235,200}

\newcommand{\cmark}{\textcolor{green!60!black}{\ding{51}}}      
\newcommand{\xmark}{\textcolor{red}{\ding{55}}} 
\title{MVP: Enhancing Video Large Language Models via Self-supervised Masked Video Prediction}

\author{
Xiaokun Sun\textsuperscript{1,2},
Zezhong Wu\textsuperscript{1,2}, 
Zewen Ding\textsuperscript{1,2},
Linli Xu\textsuperscript{1,2$\dagger$} \\
\textsuperscript{1}University of Science and Technology of China\\
\textsuperscript{2}State Key Laboratory of Cognitive Intelligence\\
{\tt\small{
\{sunxiaokun2020, felix, dingzewen\}@mail.ustc.edu.cn
}}\\
{\tt\small{linlixu@ustc.edu.cn}}
}

\begin{document}
\maketitle
\blfootnote{ $\dagger$ Corresponding author. }
\input{sections/sec0_abstract} 
\input{sections/sec1_introduction}

\input{sections/sec2_related_work}

\input{sections/sec3_method}

\input{sections/sec4_experiments}
\input{sections/sec5_conclusion}

\input{sections/sec6_Limitations}
\bibliography{custom}
\input{sections/appendix}
\end{document}

%% file: sections/sec0_abstract.tex
\begin{abstract}
Reinforcement learning based post-training paradigms for Video Large Language Models (VideoLLMs) have achieved significant success by optimizing for visual-semantic tasks such as captioning or VideoQA. 
However, while these approaches effectively enhance perception abilities, they primarily target holistic content understanding, often lacking explicit supervision for intrinsic temporal coherence and inter-frame correlations. 
This tendency limits the models' ability to capture intricate dynamics and fine-grained visual causality. 
To explicitly bridge this gap, we propose a novel post-training objective: Masked Video Prediction (MVP). 
By requiring the model to reconstruct a masked continuous segment from a set of challenging distractors, MVP forces the model to attend to the sequential logic and temporal context of events. 
To support scalable training, we introduce a scalable data synthesis pipeline capable of transforming arbitrary video corpora into MVP training samples, and further employ Group Relative Policy Optimization (GRPO) with a fine-grained reward function to enhance the model's understanding of video context and temporal properties. 
Comprehensive evaluations demonstrate that MVP enhances video reasoning capabilities by directly reinforcing temporal reasoning and causal understanding.
\end{abstract}

%% file: sections/sec1_introduction.tex
\section{Introduction}
\label{sec:intro}
Reinforcement learning with Verifiable Rewards (RLVR) has significantly enhanced the reasoning capabilities of Multimodal Large Language Models (MLLMs)~\citep{wu2025visual,xing2025caprl,feng2025video}. 
Motivated by this success, recent research has increasingly attempted to transfer RL paradigms to Video Large Language Models (VideoLLMs)~\citep{yan2025videochat,park2025deepvideo,fu2025love}. 
Currently, the approaches in the video domain involves leveraging algorithms like Group Relative Policy Optimization (GRPO)~\citep{shao2024deepseekmath} paired with diverse reward functions. 
These methods typically conduct training on standard tasks such as video question answering (videoQA)~\citep{fu2025video,hu2025videommmuevaluatingknowledgeacquisition}, video captioning~\citep{zhang2025vcapsbench}, and video grounding~\citep{gao2017tall,lei2021qvhighlightsdetectingmomentshighlights} to bolster the models' capabilities.

\begin{figure}[t]
\centering
\includegraphics[width=1.0\linewidth]{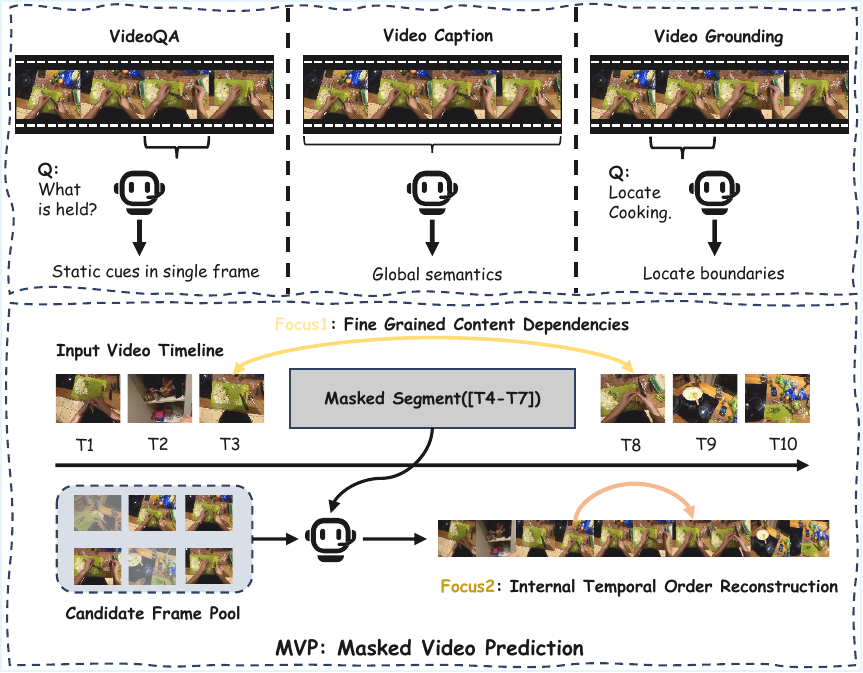}
\vspace{-18pt}
\caption{\textbf{Comparison between MVP and other tasks}. MVP compels the model to attend to both contextual content information within the video and the temporal relationships between frames.}
\label{fig:intro}
\vspace{-12pt}
\end{figure}

Despite these improvements, a critical limitation persists: most existing methods lack explicit supervision for understanding video temporal properties and inter-frame correlations. 
This deficiency hinders the models from fully grasping the intrinsic temporal nature and frame-to-frame relationships within videos. 
Moreover, obtaining high-quality training data that effectively captures video dynamics is challenging. 
Prevalent training tasks, such as videoQA, video captioning and video grounding tend to prioritize object perception, largely neglecting the \textbf{dynamic temporal properties} and the \textbf{intricate dependencies} between frames.

To address these limitations, we propose a novel post-training objective: \textbf{Masked Video Prediction (MVP)}. 
Analogous to masked token prediction in BERT~\citep{devlin2019bert} like models, MVP requires the model to reconstruct a masked video segment, compelling it to explicitly attend to sequential logic and temporal context. 
We formulate this task as a multiple-choice problem where a continuous sequence of frames is masked and mixed with intra-video distractors. 
The MVP task essentially consists of two relatively independent subtasks: frame selection and temporal ordering.
The model must select the correct frames and arrange them in the precise chronological order to fill the gap. 
Given the significant distributional shift between this task and standard pre-training data, direct Supervised Fine-Tuning (SFT) risks degrading the model's prior knowledge. 
Consequently, we adopt a Reinforcement Learning approach, utilizing Group Relative Policy Optimization (GRPO) with a fine-grained reward mechanism. 
Crucially, rather than relying on a simple binary accuracy metric, our reward structure differentiates between correct frame selection and correct temporal ordering. 
This dual-reward strategy incentivizes the model to simultaneously refine its understanding of both video content and intrinsic temporal properties. 
Additionally, we introduce a scalable MVP data synthesis pipeline capable of transforming arbitrary video corpora into training samples, enabling the generation of virtually unlimited data to meet large-scale training demands.

To rigorously validate the effectiveness and generalizability of the Masked Video Prediction task, we conduct comprehensive experiments across multiple base models and investigate the impact of data scaling by training with varying sample sizes. 
Our evaluation spanned six diverse and challenging benchmarks: VideoMME~\citep{fu2025video}, LongVideoBench~\citep{wu2024longvideobench}, LVBench~\citep{wang2025lvbench}, MLVU~\citep{zhou2024mlvu}, Video-Holmes~\citep{cheng2025video}, and TempCompass~\citep{liu2024tempcompass}. 
Empirical results consistently demonstrate that the MVP objective significantly enhances the models' video reasoning capabilities, proving its efficacy in reinforcing temporal reasoning and contextual grasp.

In summary, our main contributions are as follows:
\begin{itemize}
    \item We propose \textbf{Masked Video Prediction}, a novel post-training objective that compels VideoLLMs to master temporal logic, accompanied by a scalable data synthesis pipeline that generates unlimited training samples from arbitrary video corpora.
    \item We design a \textbf{fine-grained visual supervision reward function} within the GRPO framework. This mechanism distinguishes between frame selection and temporal ordering, providing precise feedback that ensures stable and effective model training.
    \item Extensive experiments across multiple base models and six benchmarks demonstrate that MVP significantly enhances video reasoning capabilities, proving the effectiveness and generalizability of our approach.
\end{itemize}

%% file: sections/sec2_related_work.tex
\section{Related Work}
\label{sec:related_work}

\begin{figure*}[t]
\centering
\includegraphics[width=1.0\linewidth]{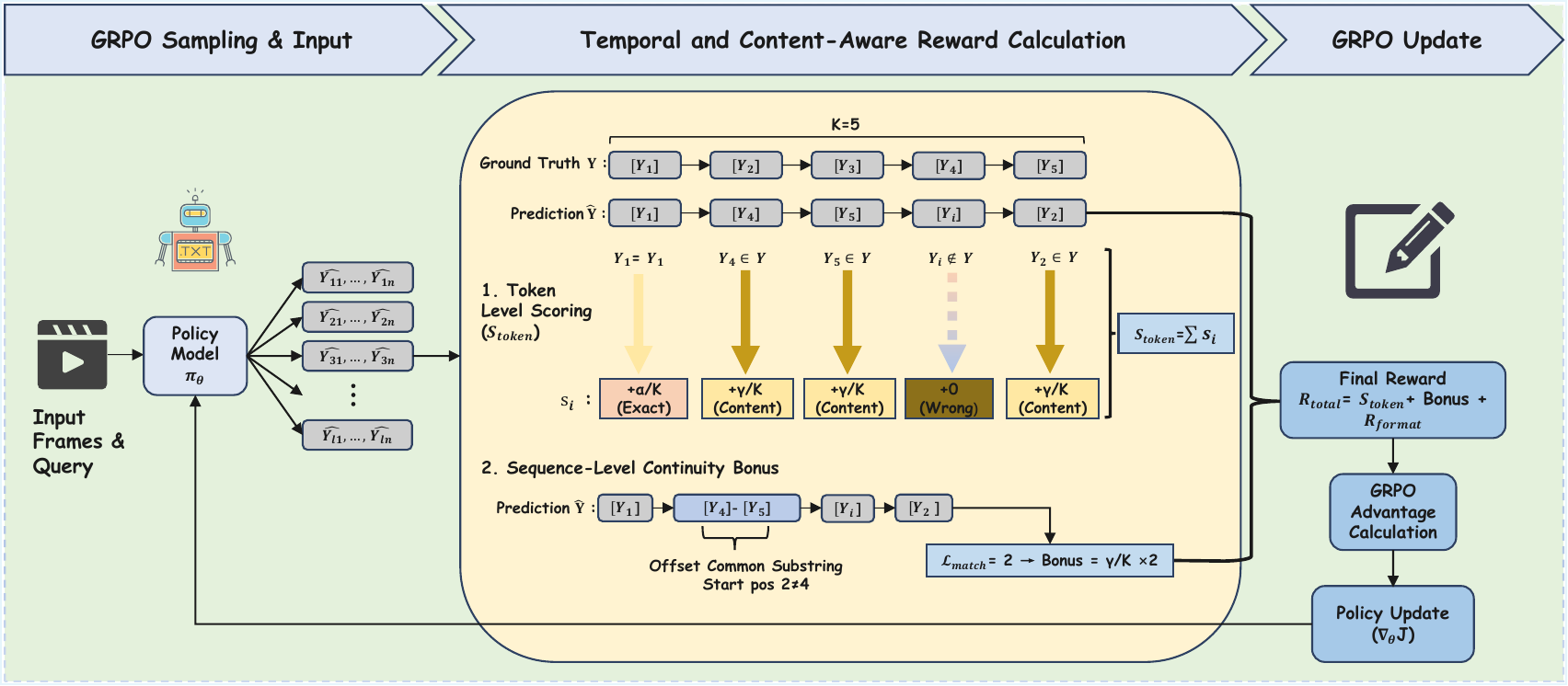}
\vspace{-10pt}
\caption{Illustration of GRPO with temporal and content-aware rewards.}
\label{fig:method}
\vspace{-10pt}
\end{figure*}

\subsection{Video Large Language Models}
Video Large Language Models (VideoLLMs) have emerged as powerful tools for video understanding~\citep{bai2025qwen3vltechnicalreport,wang2025internvl35advancingopensourcemultimodal,zhang2025llavavideovideoinstructiontuning,an2025llava,vteam2025glm45vglm41vthinkingversatilemultimodal,guo2025seed15vltechnicalreport,wang2025fostering,yang2025kwai}, building upon the strong abilities of Large Language Models (LLMs). 
These models have achieved impressive performance in tasks like video question-answering~\citep{fu2025video,zhao2025mmvu,hu2025video} and captioning~\citep{zhang2025vcapsbench} by enabling comprehensive content interpretation. 
Representative works have introduced various mechanisms to enhance this capability: BOLT~\citep{liu2025bolt} normalizes the similarity
scores between video frames and the input question into a probability distribution, and then selects frames via inverse transform sampling to reduce the input length.
SlowFast-LLaVA~\citep{xu2025slowfast} fits the two-stream SlowFast mechanism into a streamlined training pipeline to minimize input token usage while maintaining the completeness of video information. 
These advancements highlight the potential of MLLMs in advancing video understanding.

\subsection{Video Reinforcement Learning}
Recent breakthroughs by OpenAI-o1~\citep{openai2024openaio1card} and DeepSeek-R1~\citep{deepseekai2025deepseekr1incentivizingreasoningcapability} have highlighted the efficacy of Reinforcement Learning (RL) in elevating the reasoning capabilities of Large Language Models (LLMs). 
Following this trend, RL techniques such as DPO~\citep{rafailov2023direct} and GRPO~\citep{shao2024deepseekmath} have been adapted for MLLMs and VideoLLMs to enhance visual reasoning through verifiable reward mechanisms~\citep{yan2025videochat,feng2025video,xing2025caprl,park2025deepvideo,wu2025visual,tao2025digdifferentialgroundingenhancing,he2025videossrvideoselfsupervisedreinforcement,wang2025pixelreasonerincentivizingpixelspace}.
Video-R1~\citep{feng2025video} employs GRPO to improve implicit temporal and spatial reasoning by rewarding the model to utilize temporal information in vidoes.
TSPO~\cite{tang2025tspo} trains an event-aware temporal agent through reinforcement learning to effectively select frames, thereby enhancing the model's understanding of long videos.
Despite achieving a certain degree of effectiveness, these methods overlook the supervision required for the model to understand fine-grained temporal properties and inter-frame correlations in videos.

%% file: sections/sec3_method.tex
\section{Method}
\label{sec:method}

In this section, we present the proposed approach in detail. 
We first provide a formal formulation of the Masked Video Prediction (MVP) task. 
Next, we describe the scalable data synthesis pipeline designed to transform arbitrary video sources into high-quality training samples, as illustrated in Fig.~\ref{fig:datapipeline}. 
Finally, we detail the reinforcement learning framework (Fig.~\ref{fig:method}), specifically focusing on the design of fine-grained reward functions tailored to capture temporal logic and sequential dependencies.

\subsection{Masked Video Prediction}
The Masked Video Prediction (MVP) task can be conceptualized as a "visual cloze test" for videos. 
It requires the model to reconstruct a missing video segment by selecting the correct frames from a candidate pool and arranging them in the correct chronological order based on the surrounding context.
Formally, given a video sequence $V = \{f_1, f_2, ..., f_N\}$, a continuous segment $V_{target}$ is masked, leaving the observable context $V_{context} = V \setminus V_{target}$. 
The masked segment is decomposed into $K$ ordered key frames, forming the positive set $\mathcal{S}_{pos} = \{p_1, ..., p_K\}$. 
These are mixed with a set of hard negative distractors $\mathcal{S}_{neg}$ sampled from the same video to form a shuffled candidate pool $\mathcal{C} = \text{Shuffle}(\mathcal{S}_{pos} \cup \mathcal{S}_{neg})$. 
The objective is to predict a sequence of indices $Y = \{y_1, ..., y_K\}$ that selects the correct frames from $\mathcal{C}$ and arranges them in their original temporal order, such that the reconstructed sequence logically bridges the gap in $V_{context}$.

\subsection{Scalable Data Synthesis Pipeline}

\begin{figure}[t]
  \centering
   \setlength{\abovecaptionskip}{0.cm}
   \setlength{\belowcaptionskip}{0.cm}
   \includegraphics[width=\linewidth]{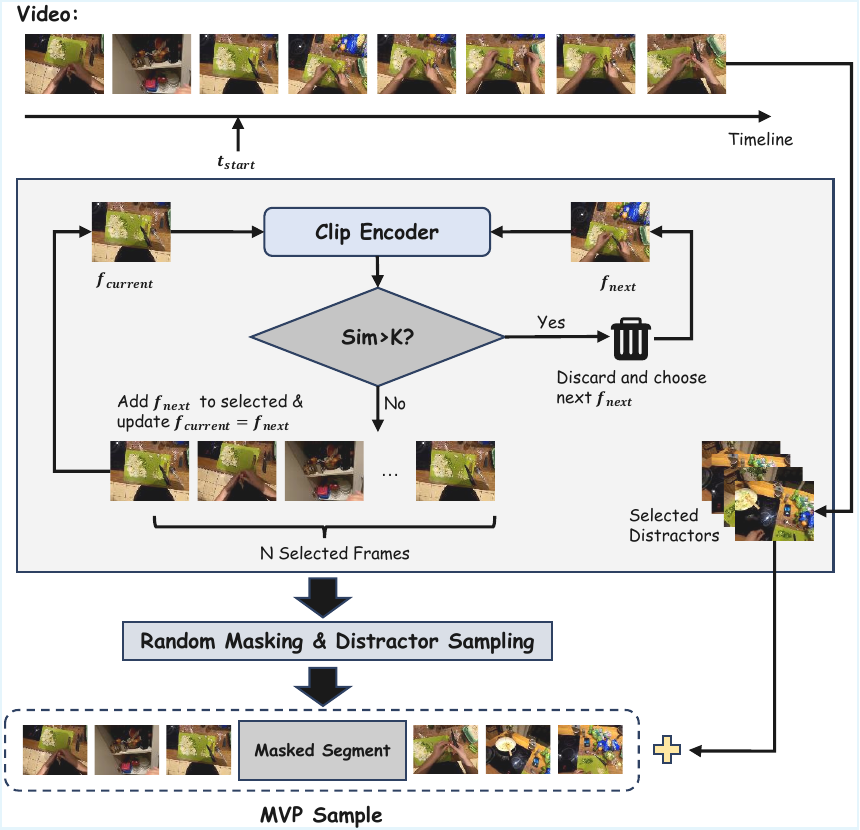}
   \caption{\textbf{Scalable Data Synthesis Pipeline.} We aim to ensure that the MVP samples contain minimal redundancy, enabling the model trained on this data to effectively learn temporal relationships in videos and dependencies between frames.}
   \label{fig:datapipeline}
\end{figure}

To systematically investigate the impact of MVP on video understanding capabilities, we design a scalable data synthesis pipeline capable of generating MVP training samples from arbitrary video sources as shown in figure~\ref{fig:datapipeline}. 
Our objective is to mask semantically significant segments within continuous video streams, thereby necessitating the model to capture underlying temporal logic and long-range sequential dependencies.

Given the inherent redundancy in video data, where consecutive frames often exhibit high visual similarity, a naive frame selection strategy would result in information-poor samples. 
To mitigate this, we employ a visual de-duplication strategy. 
For a given video, we first load frames at 1 FPS. Starting from a randomly selected timestamp $t_{start}$, we aim to select a sequence of $N$ distinct frames. 
Let $f_{current}$ denote the most recently selected frame. 
We iterate through subsequent frames $f_{next}$ and compute their visual similarity using a CLIP~\citep{radford2021learning} encoder $\phi(\cdot)$:
\vspace{-5pt}
\begin{equation}
    s = \phi(f_{curr}) \cdot \phi(f_{next})^\top
    \label{eq:clip_sim}
\end{equation}

\vspace{-5pt}
If $s > \kappa$ (where $\kappa$ is a pre-defined threshold), $f_{next}$ is discarded as redundant. 
We continue this process until we find a frame where the similarity falls below $\kappa$, at which point it is added to our selected set, and the process repeats until $N$ frames are collected.

From this selected sequence of $N$ frames, we randomly mask $m$ frames to serve as the prediction targets. 
To construct a challenging candidate pool, we randomly sample $l$ distractor frames from the temporal vicinity (before or after) of the selected $N$-frame segment within the same video. 
These distractors are mixed with the $m$ target frames to form the final shuffled candidate set.

This pipeline is universally applicable to any video corpus, allowing for the effortless generation of large-scale datasets. 
Furthermore, by varying the starting timestamp and random masking, a single video can yield multiple distinct training samples, maximizing data utilization and providing diverse temporal contexts

\subsection{GRPO with Temporal and Content-Aware Rewards}

\subsubsection{Preliminary: GRPO}

Group Relative Policy Optimization (GRPO) operates as a highly efficient reinforcement learning framework that explicitly eliminates the need for a separate value function critic, thereby significantly reducing both memory usage and computational overhead. 
For each input query $q$, GRPO samples a cohort of $G$ outputs $\{o_1, \dots, o_G\}$ from the current policy $\pi_{\theta_{\text{old}}}$. 
The advantage $A_i$ for each individual output $o_i$ is then derived by normalizing its reward $r_i$ using the mean and standard deviation of the rewards within that group:

\begin{equation}
    A_i = \frac{r_i - \mathrm{mean}(\{r_j\})}{\mathrm{std}(\{r_j\}) + \epsilon}
    \label{eq:advantage}
\end{equation}

\noindent The objective function maximizes the clipped surrogate objective, regularized by a KL-divergence term:
\begin{equation}
    \mathcal{J} = \frac{1}{G} \sum_{i=1}^G \left( \mathcal{L}^{clip}_i - \beta D_{KL}(\pi_\theta || \pi_{ref}) \right)
    \label{eq:grpo}
\end{equation}

\noindent where $\mathcal{L}^{clip}_i = \min(\rho_i A_i, \mathrm{clip}(\rho_i, 1\pm\epsilon)A_i)$ and $\rho_i = \pi_\theta(o_i|q)/\pi_{\theta_{old}}(o_i|q)$.

\subsubsection{Reward Design}
To precisely guide the model in mastering both visual content identification and temporal reasoning, we design a hierarchical reward function. Given the ground truth sequence $Y = \{y_1, \dots, y_K\}$ and the predicted sequence $\hat{Y} = \{\hat{y}_1, \dots, \hat{y}_K\}$, the total correctness reward $R_{correct}$ is composed of token-level matching scores and sequence-level continuity bonuses.

\paragraph{Token-Level Scoring.}
First, we evaluate each predicted item $\hat{y}_i$ at position $i$ to determine if the model has correctly identified the content and its temporal placement. 
The scoring function $s(i)$ assigns full credit for exact matches (content + position) and partial credit for content retrieval without correct ordering:
\begin{equation}
    s(i) = 
    \begin{cases} 
        \alpha/K, & \hat{y}_i = y_i \\
        \gamma/K, & \hat{y}_i \in Y \setminus \{y_i\} \\
        0, & \text{otherwise}
    \end{cases}
    \label{eq:token_score}
\end{equation}
Here, we set $\alpha > \gamma$ to strongly incentivize correct temporal ordering, and $\gamma$ acknowledges the model's ability to recognize valid visual content even if the timestamp is misplaced. 
The total token score is $S_{token} = \sum_{i=1}^K s(i)$.

\paragraph{Sequence-Level Continuity Bonus.}
To further enforce temporal logic, we detect the \textbf{Common Substrings} (length $\ge 2$) between $\hat{Y}$ and $Y$. 
We apply a continuity bonus mechanism that rewards preserved temporal substructures. 
Specifically, for all matching substrings that do not start at the correct absolute position (i.e., offset matches), we provide an additional reward. 
This encourages the model to capture relative temporal order even when global alignment is imperfect. 
Let $\mathcal{L}_{match}$ be the sum of lengths of such valid substrings. 
The final reward combines the token-level accuracy with this structural bonus:
\begin{equation}
    R_{correct} = S_{token} + \frac{\gamma}{K} \times \mathcal{L}_{match}
    \label{eq:total_score}
\end{equation}

Through this granular reward design, we encourage the model to incrementally learn both content identification and temporal sequencing, ultimately fostering more robust video reasoning skills.
\begin{table*}[t]
\centering

\small
\setlength{\tabcolsep}{3mm}
\resizebox{0.95\textwidth}{!}{
\begin{tabular}{lcccccccc}
\toprule
\textbf{Model} &\textbf{Frames}& \textbf{VideoMME} & \textbf{LongVideoBench} & \textbf{LVBench} & \textbf{MLVU} & \textbf{Video-Holmes} & \textbf{TempCompass}  \\
\midrule
\textcolor{graytext}{Qwen2.5-VL-72B}  & 768 & 73.3 & 60.7 & 47.3   & 74.6 & --   & 74.8 \\
\textcolor{graytext}{Qwen3-VL-235B-A22B}  & 768 & 79.0 & -- & 63.6   & 83.8 & --   & -- \\
\textcolor{graytext}{InternVL3.5-241B-A28B} & - & 72.9 & 67.1 & -- & 78.2 & --   & --   \\
\midrule

\rowcolor{bluehighlight} & 64  & 61.2 & 53.2 & 36.1 & 61.4 & 36.9 & 69.0 \\
 \rowcolor{bluehighlight}
 \multirow{-2}{*}{Jigsaw-7B$^*$}                                      & 128 & 62.4   & 56.2   & 38.0   & 64.6   & 37.6   & --   \\

\multirow{2}{*}{Qwen2.5-VL-7B-Instruct$^*$} & 64  & 60.3 & 48.0 & 32.9 & 57.1 & 33.4 & 66.0 \\
                                        & 128 & 60.9   & 51.4   & 37.5   & 63.0   & 35.0   & --   \\
\rowcolor{greenhighlight}
                                        & 64  & 61.2{\color{red}$^{\uparrow0.9}$} & 55.9{\color{red}$^{\uparrow7.9}$} & 38.2{\color{red}$^{\uparrow5.3}$} & 62.2{\color{red}$^{\uparrow5.1}$} & 35.9{\color{red}$^{\uparrow2.5}$} & 68.5{\color{red}$^{\uparrow2.5}$} \\
\rowcolor{greenhighlight}
\multirow{-2}{*}{\quad +MVP (Ours)}     & 128 & 63.5{\color{red}$^{\uparrow2.6}$} & 58.6{\color{red}$^{\uparrow7.2}$} & 41.1{\color{red}$^{\uparrow3.6}$} & 66.0{\color{red}$^{\uparrow3.0}$} & 36.7{\color{red}$^{\uparrow1.7}$} & -- \\ 

\multirow{2}{*}{InternVL3.5-8B$^*$}         & 64  & 60.4 & 53.0 & 37.3 & 63.5 & 39.3 & 68.8 \\
                                        & 128 & 62.2   & 55.6   & 42.4   & 65.1   & 37.7   & --   \\
\rowcolor{greenhighlight}
                                        & 64  & \textbf{63.6}{\color{red}$^{\uparrow3.2}$} & 61.0{\color{red}$^{\uparrow8.0}$} & \textbf{42.4}{\color{red}$^{\uparrow5.1}$} & \textbf{67.0}{\color{red}$^{\uparrow3.5}$} & 39.4{\color{red}$^{\uparrow0.1}$} & 72.3{\color{red}$^{\uparrow3.5}$} \\
\rowcolor{greenhighlight}
\multirow{-2}{*}{\quad +MVP (Ours)}     & 128 & 64.0{\color{red}$^{\uparrow1.8}$} & 59.5{\color{red}$^{\uparrow3.9}$} & \textbf{44.2}{\color{red}$^{\uparrow1.8}$} & \textbf{69.0}{\color{red}$^{\uparrow3.9}$} & 38.8{\color{red}$^{\uparrow1.1}$} & -- \\

\multirow{2}{*}{Qwen3-VL-8B-Thinking$^*$}   & 64  & 62.6 & 59.6 & 38.1 & 59.7 & 38.5 & 74.2 \\
                                        & 128 & 67.4   & 62.9   & 41.3   & 65.4   & 41.8   & --   \\
\rowcolor{greenhighlight}
                                        & 64  & 62.7{\color{red}$^{\uparrow0.1}$} & \textbf{62.0}{\color{red}$^{\uparrow2.4}$} & 39.8{\color{red}$^{\uparrow1.7}$} & 63.6{\color{red}$^{\uparrow3.9}$} & \textbf{40.1}{\color{red}$^{\uparrow1.6}$} & \textbf{74.7}{\color{red}$^{\uparrow0.5}$} \\
\rowcolor{greenhighlight}
\multirow{-2}{*}{\quad + MVP (Ours)}    & 128 & \textbf{67.6}{\color{red}$^{\uparrow0.2}$} & \textbf{63.8}{\color{red}$^{\uparrow0.9}$} & 43.0{\color{red}$^{\uparrow1.7}$} & 67.8{\color{red}$^{\uparrow2.4}$} & \textbf{42.6}{\color{red}$^{\uparrow0.8}$} & -- \\
\bottomrule
\end{tabular}}
\caption{\textbf{Comparisons with state-of-the-art methods on various benchmarks.} We report results with 64 and 128 frames for models trained with MVP. The * denotes results evaluated in our experimental settings and the bold text means the best performance under this frame setting.}
\label{tab:main_res}
\end{table*}

\paragraph{Format Reward.}
To encourage the model to perform explicit reasoning before generating the final answer, we introduce a format-based reward that constrains the structure of the model output. 
Specifically, the model is required to enclose its intermediate reasoning process within \texttt{<think>} tags and to present the final prediction within \texttt{<answer>} tags. 
Based on this requirement, the format reward $R_{\text{format}}$ is defined as an indicator function that returns 1 if and only if the generated output strictly adheres to the prescribed structure, and 0 otherwise. 
By explicitly enforcing this structural constraint, the model is compelled to externalize its reasoning process prior to producing an answer, thereby facilitating a clear and interpretable \textit{think-before-answer} mechanism.

We formulate the final training objective as a composite reward function that jointly accounts for both formatting compliance and task-specific correctness. The total reward is defined as a weighted sum of the format reward and the correctness reward:
\begin{equation}
    R_{\text{total}} = \beta R_{\text{format}} + (1 - \beta) R_{\text{correct}},
    \label{eq:total_reward}
\end{equation}
where $\beta \in [0,1]$ is a balancing coefficient that controls the relative importance of enforcing the reasoning structure versus optimizing task performance. 
This compound reward formulation jointly promotes task accuracy and structural compliance, resulting in outputs that are both accurate and interpretable.

\subsubsection{Policy Optimization}
We integrate the proposed fine-grained reward function $R_{total}$ into the GRPO framework to optimize the VideoLLM policy $\pi_\theta$. 
For each masked video query $q$, the model generates a group of $G$ candidate sequences $\{\hat{Y}_1, \dots, \hat{Y}_G\}$ from the current policy $\pi_{\theta_{old}}$. 
We compute the reward $r_i = R_{total}(\hat{Y}_i, Y)$ for each candidate using Eq~\ref{eq:total_reward}, which explicitly values both frame identification and temporal sequencing.

The advantages $A_i$ are then derived using the group statistics as defined in Eq~\ref{eq:advantage}.
Finally, the policy is updated by maximizing the objective $\mathcal{J}$ in Eq~\ref{eq:grpo}. 
By directly optimizing for $R_{total}$ within this group-relative formulation, the model is incentivized to internalize the underlying temporal logic, effectively aligning its generation probabilities with temporally coherent video reasoning pathways without the computational overhead of a value network.

%% file: sections/sec4_experiments.tex
\section{Experiments}
\label{sec:exper}

\begin{figure*}[t]
\centering
\includegraphics[width=1.0\linewidth]{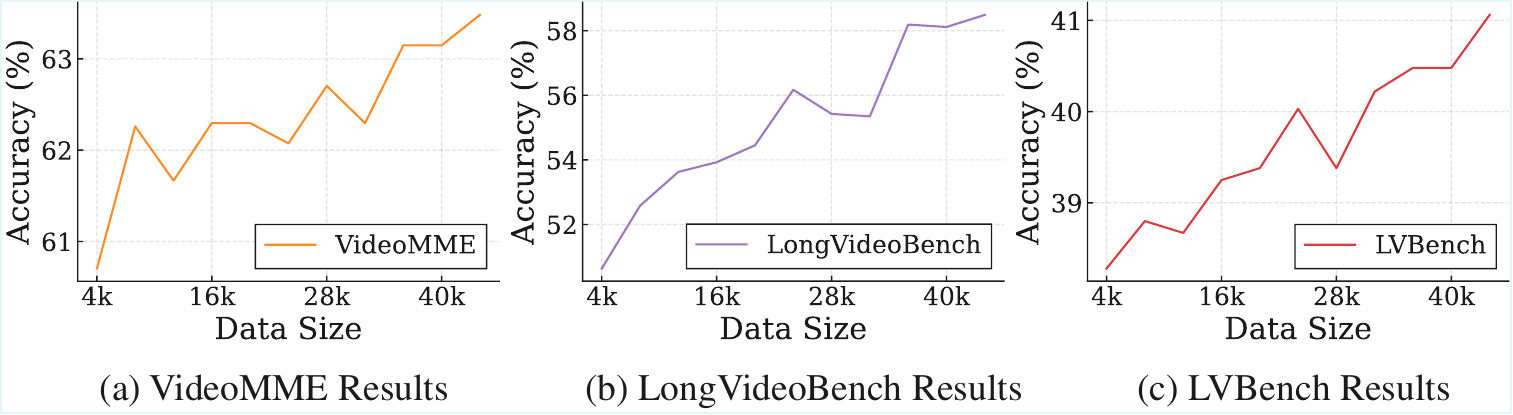}
\vspace{-10pt}
\caption{\textbf{Data scaling analysis on different benchmarks.} We train for 1 epoch on different data sizes and evaluate the results on three different benchmarks.}
\label{fig:data}
\vspace{-10pt}
\end{figure*}
\subsection{Experimental Settings}
\subsubsection{Benchmarks}
We conduct a comprehensive evaluation of MVP across six video benchmarks: VideoMME~\citep{fu2025video}, LongVideoBench~\citep{wu2024longvideobench}, LVBench~\citep{wang2025lvbench}, MLVU~\citep{zhou2024mlvu}, Video-Holmes~\citep{cheng2025video} and TempCompass~\citep{liu2024tempcompass}. 
These benchmarks were selected to cover a broad spectrum of video reasoning capabilities. 
VideoMME and MLVU primarily assess general VideoQA performance.
VideoMME comprises videos spanning diverse themes and categorized into short, medium, and long durations. 
Each video is paired with three questions, enabling a comprehensive assessment of the model's video question-answering capabilities. 
MLVU consists of nine subtasks, designed to evaluate model performance from multiple distinct perspectives.
LVBench and LongVideoBench focus on long-video understanding, inlcuding extremly long videos ranging up to two hours. 
Video-Holmes serves as a reasoning benchmark, specifically targeting complex video reasoning abilities, while TempCompass evaluates the model's proficiency in temporal understanding.

\subsubsection{Implementation Details.}
We conduct GRPO training for MVP on three base models: Qwen2.5-VL-7B-Instruct~\citep{bai2025qwen25vltechnicalreport}, InternVL3.5-8B~\citep{wang2025internvl35advancingopensourcemultimodal}, and Qwen3-VL-8B-Thinking~\citep{bai2025qwen3vltechnicalreport}. 
The training utilizes 50k data samples synthesized from the LLaVA-Video-178K dataset~\citep{zhang2025llavavideovideoinstructiontuning}. 
Each MVP training sample is constructed with a sequence length of 15 frames. 
Preliminary tests(included in the appendix~\ref{apx:MVP}) indicate that the task presents a significant challenge to the models, therefore, we avoid masking an excessive number of frames. 
The final configuration consists of 10k samples with 2 masked positions, 25k samples with 3 masked positions and 15k samples with 4 masked positions. 
All models are trained under the same MVP formulation and reward design to ensure a fair comparison across architectures.
For evaluation, we test each dataset using both 64-frame and 128-frame settings (except of TempCompass which include mostly short videos, so we just test it under 64-frame settings), with the token count per frame limited to 256.
Detailed hyperparameters for the data synthesis process and training, along with prompts and other settings are provided in the appendix~\ref{apx:imp_detail}. 

\subsection{Experiment Results}
\subsubsection{Results on Standard Benchmarks.}
As presented in Table \ref{tab:main_res}, MVP yields comprehensive and consistent improvements across a broad spectrum of video understanding domains, ranging from general perception and long-video understanding to complex reasoning and temporal logic. 
Crucially, these gains are observed uniformly across diverse base models, regardless of their specific architecture or training paradigm—spanning standard instruction-tuned models like Qwen2.5-VL-7B-Instruct, as well as reasoning-oriented models like Qwen3-VL-8B-Thinking and InternVL3.5-8B. For instance, we observe significant enhancements in long-context tasks (e.g., substantial improvements on LongVideoBench and LVBench across all backbones), alongside robust boosts in general QA (VideoMME, MLVU) and reasoning-intensive benchmarks (Video-Holmes, TempCompass). 
This universality underscores that MVP provides a fundamental visual supervision signal that effectively generalizes across different model types.
To benchmark our approach against existing pre-training objectives, we implemented Jigsaw-7B~\citep{wu2025visual} as a baseline, which is trained to reorder shuffled video segments using 100k samples synthesized from the same LLaVA-Video-178K dataset. 
Even with using only half the size of training data of Jigsaw-7B, our MVP method consistently delivers better results across most benchmarks, leading by margins such as 2.7 points on LongVideoBench.
This demonstrates that while segment reordering provides only coarse-grained supervision, the MVP task compels the model to master fine-grained inter-frame relationships and intrinsic temporal properties, resulting in a more robust video representation.
Notably, these enhancements persist across different input frame settings (64 and 128). This consistency demonstrates that the efficacy of MVP is independent of specific sampling densities, suggesting that the model acquires fundamental temporal reasoning skills that remain effective across varying temporal resolutions.

\subsubsection{Ablation Study and Analysis.}

\begin{table*}[t]
\centering

\small
\setlength{\tabcolsep}{3mm}
\resizebox{0.95\textwidth}{!}{
\begin{tabular}{lcccccccc}
\toprule
\textbf{Content}&\textbf{Sequence}& \textbf{VideoMME} & \textbf{LongVideoBench} & \textbf{LVBench} & \textbf{MLVU} & \textbf{Video-Holmes} & \textbf{TempCompass}  \\
\midrule
  \xmark& \xmark & 60.4 & 53.5 & 37.0   & 60.5 & 34.0   & 67.1 \\
  \cmark& \xmark & 61.0 & 55.5 &  37.4  & 61.3 & 35.1   & 68.1 \\
\cmark & \cmark & \textbf{61.2} & \textbf{55.9} & \textbf{38.2} & \textbf{62.2} & \textbf{35.9}   & \textbf{68.5}   \\

\bottomrule

\end{tabular}}
\caption{\textbf{Ablation study on different reward components.} Content means content rewards, without which we only give a reward when the model predicts an exact match (position and content are all correct). Sequence means sequence-level continuity bonus. All experiments are conducted using 64 frames.}
\label{tab:ablation}
\end{table*}

\paragraph{Ablation on Reward Components.} To validate the effectiveness of our fine-grained reward mechanism, we dissect the impact of each component in Table \ref{tab:ablation}. 
We train the Qwen2.5-VL-7B-Instruct backbone under various reward settings. 
To ensure a rigorous comparison, all models are evaluated using checkpoints obtained at an identical number of training steps.
We establish a baseline using a strict Exact Match policy (Row 1), where the model receives rewards only when both the frame content and its temporal index are perfectly predicted, with no partial credit for misplaced frames. 
Introducing the Content-Aware component (Row 2), which grants partial rewards for correctly identifying target frames regardless of their order, leads to immediate performance gains (e.g., +2.0 on LongVideoBench). 
This indicates that encouraging the model to distinguish relevant visual information from distractors is a crucial first step. 
Finally, incorporating the Sequence-Level Continuity Bonus (Row 3) yields the best performance across all benchmarks, which demonstrates that explicitly incentivizing the preservation of temporal substructures is essential for the model to master the intrinsic sequential logic and causal dynamics of videos.

\paragraph{Data Scaling Analysis.} To validate the scalability of our approach, we investigate the impact of data scaling by training the Qwen2.5-VL-7B model on MVP subsets of varying sizes. 
We adhered to a rigorous experimental protocol: all models are trained for a single epoch with a fixed mask ratio (2:3:4 frames = 1:2:1) to isolate the effect of data volume. 
As shown in Figure~\ref{fig:data}, the results on VideoMME, LongVideoBench and LVBench demonstrate a monotonic increase in performance as the dataset size grows. 
Crucially, the performance curve shows no signs of saturation within our tested range, indicating that the MVP objective effectively leverages additional data to refine the model's temporal reasoning capabilities. 
This trend confirms the scalable nature of the MVP task. 
Moreover, given the model's responsiveness to data quantity, we posit that scaling up the quality and diversity of the training corpus would likely unlock further improvements in video reasoning.

\paragraph{Training Curves.}

\begin{figure}[t]
  \centering
   \setlength{\abovecaptionskip}{0.cm}
   \setlength{\belowcaptionskip}{-10pt}
   \includegraphics[width=\linewidth]{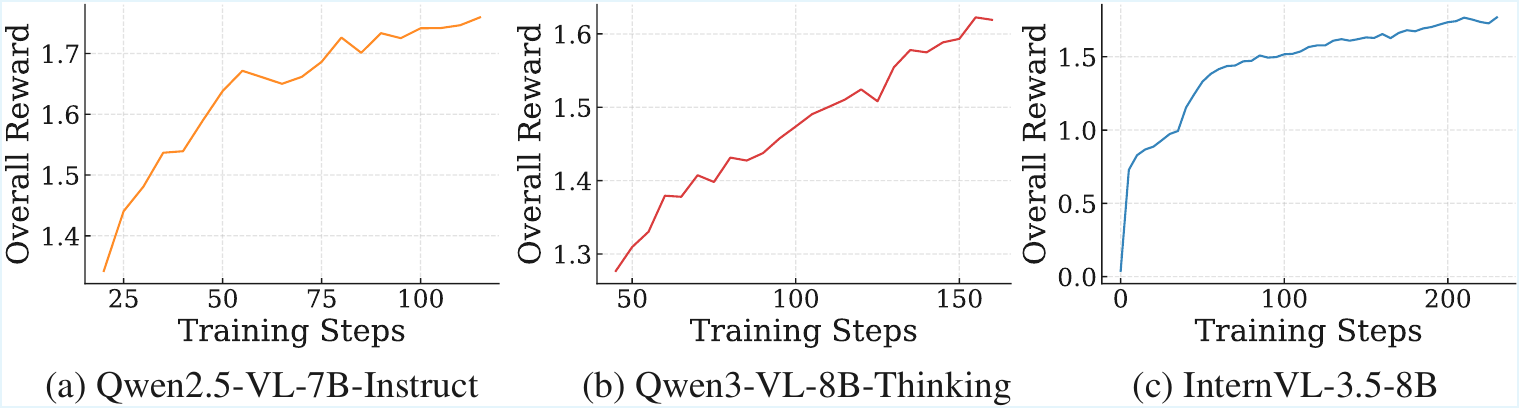}
   \caption{Training curves on different backbones.}
   \label{fig:curves}
\end{figure}

To better illustrate the optimization dynamics, we visualize the training process by plotting the overall validation reward curves for all three backbone models. 
As shown in Figure~\ref{fig:curves}, the reward values consistently increase over training iterations across all architectures, indicating smooth and stable learning behavior. 
This steady improvement provides strong empirical evidence for the effectiveness of our training strategy, demonstrating that the proposed fine-grained reward design reliably guides the models toward better frame selection and temporal ordering capabilities. 
Notably, the absence of oscillation or collapse suggests that the training process avoids common reinforcement learning instabilities. 
Moreover, the similar convergence trends observed across different backbones indicate that the proposed framework generalizes well across architectures, underscoring the robustness and model-agnostic nature of the MVP training paradigm.

\paragraph{Qualitative Analysis.}

\begin{figure*}[t]
\centering
\includegraphics[width=1.0\linewidth]{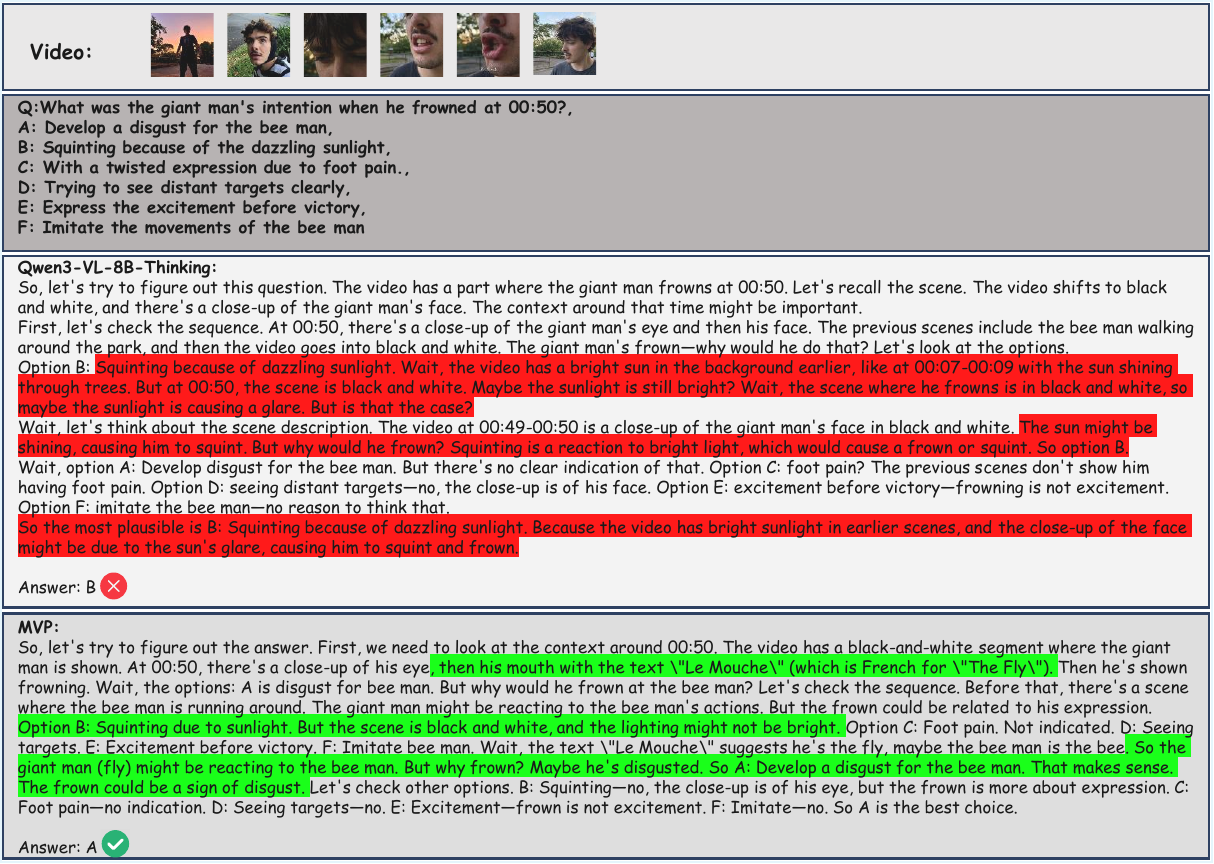}
\vspace{-10pt}
\caption{\textbf{An example of the reasoning process after training with MVP.} MVP training enhances the model's grasp of video temporality and focuses it more on visual cues to reduce hallucinations.}
\label{fig:case}
\vspace{-10pt}
\end{figure*}

To qualitatively illustrate MVP's impact on fine-grained perception and inter-frame reasoning, Figure~\ref{fig:case} compares the reasoning trajectories of the base and MVP-aligned models. 
As shown, the base model is prone to temporal hallucinations driven by spurious correlations; it often incorrectly propagates past context (e.g., environmental conditions) to explain subsequent events, resulting in plausible but factually incorrect physical interpretations. 
In contrast, the MVP model demonstrates superior grounding capabilities. 
By learning to verify temporal continuity through masked prediction, it effectively overrides such hallucinatory priors, instead anchoring its reasoning in subtle but critical visual details present in the relevant frames. 
The visualization confirms that MVP enables the model to correctly interpret complex causal dynamics, such as distinguishing emotional reactions from physical reflexes, by strictly adhering to the verified visual evidence rather than relying on static semantic associations.
This capability reveals a robust verification mechanism, effectively transforming video reasoning from passive pattern matching into active, evidence-based inquiry.

%% file: sections/sec5_conclusion.tex
\section{Conclusion}
\label{sec:conclusion}
\vspace{-0.8em}
In this work, we address the lack of explicit temporal supervision in VideoLLMs by introducing MVP, a novel post-training objective that encourages models to learn intrinsic temporal dynamics and fine-grained inter-frame relationships.
Leveraging a scalable data synthesis pipeline and a hierarchical reward design within the GRPO framework, MVP effectively converts arbitrary video data into high-quality supervision for temporal reasoning. 
Extensive evaluations across multiple benchmarks show that MVP consistently improves performance on diverse video reasoning tasks. 
Moreover, we observe a monotonic performance gain with increased data scale, underscoring the scalability of the approach and its potential as a strong foundation for future video reasoning models.

%% file: sections/sec6_Limitations.tex
\section*{Limitations}
\label{sec:limitations}
Although MVP demonstrates significant efficacy in enhancing video reasoning, we acknowledge certain limitations in the current scope of our work. 
First, while the MVP task necessitates implicit temporal logic, it lacks direct supervision on the explicit reasoning process. 
The model is trained to optimize the final selection and ordering outcomes, but is not explicitly guided to verbalize the underlying causal relationships or logical deductions behind its decisions.
Future research will explore integrating process-level supervision to the model's reasoning trajectory, fostering a more robust and transparent understanding of video content and temporal relationships.

Second, due to time and computational resource constraints, we have not fully explored the asymptotic limits of the MVP task. 
Our current experiments rely on a relatively small amount of training data and limited training duration, meaning the model's performance has not yet truly converged. 
As indicated by our scaling analysis, performance continues to improve with increased data, suggesting that the task's full potential is yet to be realized. 
Investigating the performance upper bounds through large-scale training remains an important direction for future work.

\section*{Ethical Considerations}
\paragraph{Data Privacy and Consent.} 
Our study exclusively uses publicly available datasets. 
These datasets were released under open-source licenses by their original curators, ensuring that the data collection process adhered to ethical standards. 
We have not attempted to re-identify any individuals or extract personally identifiable information from the video data. 
\paragraph{Ethical Use of Models.}
The proposed Masked Video Prediction (MVP) framework is designed for research purposes to enhance Video LLMs. 
We have conducted manual spot-checks on model outputs to ensure they do not generate offensive or biased content. 
\paragraph{Environmental Impact.} 
To reduce the carbon footprint, we prioritize energy-efficient self-supervised learning objectives and utilize pre-trained models whenever possible to minimize redundant computational costs.

%% file: sections/appendix.tex
\appendix

\section{Use of AI Assistants}

We acknowledge the use of AI assistants in the preparation of this work. Their involvement was strictly confined to the following aspects:

\begin{itemize}
    \item \textbf{Language Refinement:} The AI was used to improve the grammatical flow, clarity, and conciseness of the manuscript.
    \item \textbf{Engineering Support:} The AI assisted in writing boilerplate code and debugging scripts for the experimental infrastructure.
\end{itemize}

All fundamental scientific contributions, including the conceptualization of the research, the design of the methodology, the analysis of results and the primary drafting of the manuscript, were performed entirely by the human authors. The authors retain full responsibility for the accuracy and integrity of the final paper.

\section{Reproducibility and Stability}
To ensure the reliability of our results, all experiments were repeated across three to five independent runs. 
We observed that the outcomes remained identical due to the use of fixed random seeds and a deterministic greedy decoding strategy (temperature = 0) during inference. 
Consequently, we report the consistent scores obtained across all trials.

\section{Further Implementation Details}
\label{apx:imp_detail}
\subsection{Training Details}
For the rollout process, the \texttt{rollout\_batch\_size} was configured to 256 for both Qwen2.5-VL-7B-Instruct and Qwen3-VL-8B-Thinking, while a smaller size of 128 was utilized for InternVL-3.5-8B. 
Detailed configurations for the remaining rollout parameters are provided in Table~\ref{tab:rollout}. 
Both the \texttt{max\_prompt\_length} and \texttt{max\_response\_length} were fixed at 8192 tokens. 
During the training phase, an additional 2,000 samples were reserved exclusively for validation. 
All models were evaluated after a single training epoch to maintain consistency.
For the reward function parameters, we set $\alpha$ for rewarding exact matches to 3.0, $\beta$ for balancing format and correct rewards to 0.1 and $\gamma$ for rewarding correct content selection to 0.9.

\subsection{Data Construction Details}
To identify and manage temporal redundancy within the video sequences, we set the redundancy threshold $\kappa$ to 0.95. 
For each selected sample, we supplemented the $m$ masked frames with $6-m$ distractor frames sampled from the same source video, thereby ensuring a consistent set of six candidates for the model to choose from. 

To further guarantee the quality of temporal reasoning and content representation, we implemented a rigorous filtering pipeline. 
Specifically, each candidate sample was evaluated by Qwen2.5-VL-72B-Instruct through 10 independent rollouts. 
Samples that failed to accrue any points across all rollouts---based on the reward function defined in Section~\ref{sec:method}---were deemed to lack meaningful temporal information and were subsequently excluded from the dataset.

\section{Prompts}
\begin{figure*}[t]
\centering
\includegraphics[width=1.0\linewidth]{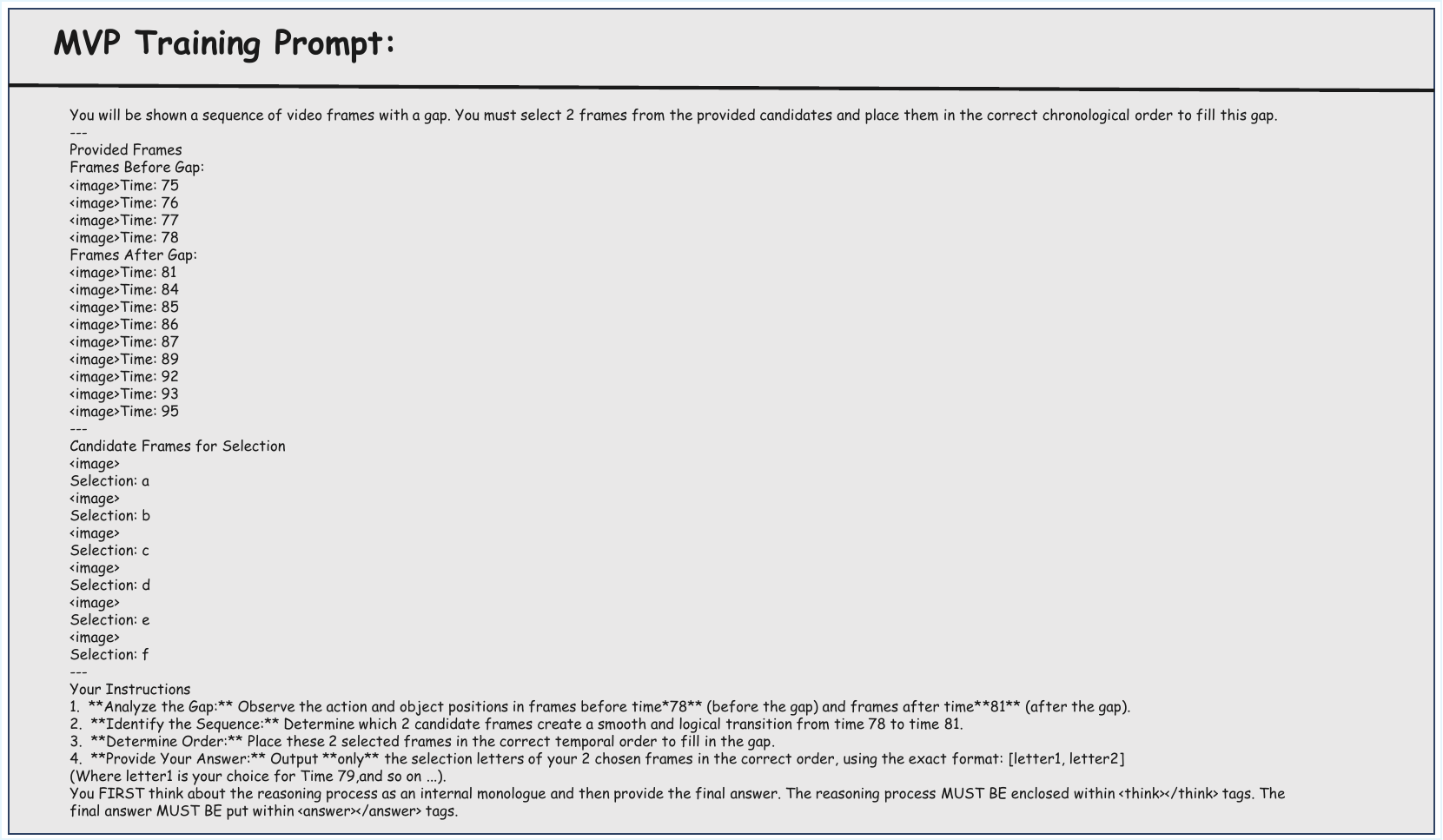}
\vspace{-10pt}
\caption{\textbf{Training prompt for MVP.}}
\label{fig:train_prompt}
\vspace{-10pt}
\end{figure*}

\begin{figure*}[t]
\centering
\includegraphics[width=1.0\linewidth]{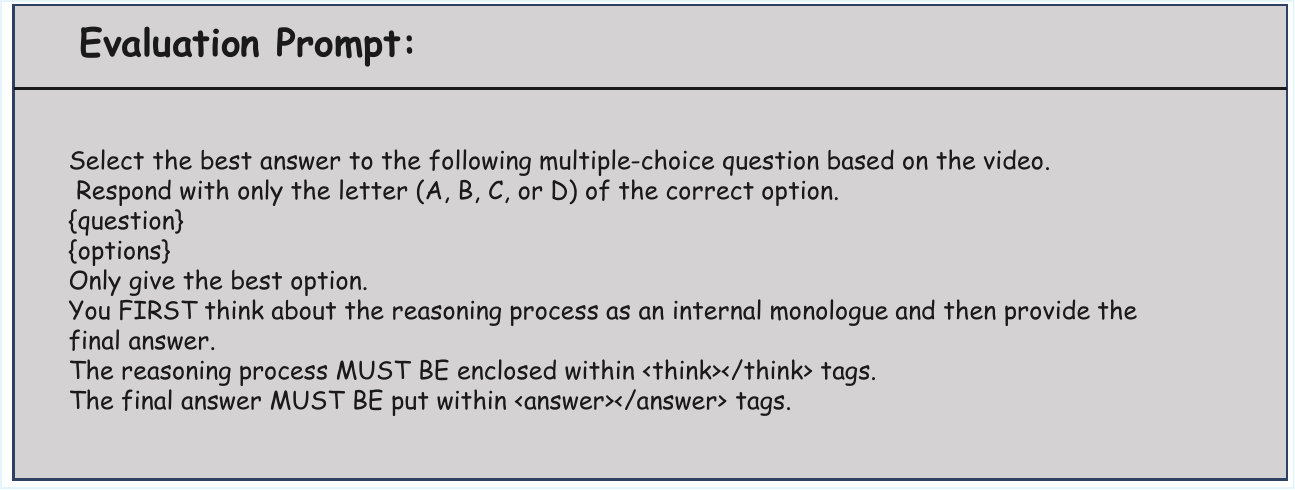}
\vspace{-10pt}
\caption{\textbf{Evaluation prompt.}}
\label{fig:test_prompt}
\vspace{-10pt}
\end{figure*}
We provide detailed prompts for the MVP training and for evaluating on diiferent benchmarks in this section, which are shown in Figure~\ref{fig:train_prompt},~\ref{fig:test_prompt}.

\begin{table}[t]
    \centering

    \footnotesize 
    \begin{tabularx}{0.7\linewidth}{lX} 
        \toprule
        \textbf{Parameter} & \textbf{Value} \\
        \midrule
        n (rollout number) & 5 \\
        temperature & 1.0 \\
        top\_p & 1.0 \\
        limit\_images & 0 \\
        gpu\_memory\_utilization & 0.6 \\
        enforce\_eager & false \\
        enable\_chunked\_prefill & false \\
        tensor\_parallel\_size & 2 \\
        \midrule
        \multicolumn{2}{l}{\textbf{Validation Configs}} \\
        -- temperature & 0.6 \\
        -- top\_p & 0.95 \\
        -- n & 1 \\
        \bottomrule
    \end{tabularx}
    \caption{Rollout Configuration.}
    \label{tab:rollout}
\end{table}

\section{Performance of Base Models on MVP}
\label{apx:MVP}

Table~\ref{tab:mvp_results} presents a comprehensive baseline evaluation of the vision-language models employed in our study, including Qwen2.5-VL-7B-Instruct, Qwen2.5-VL-8B-Instruct-Thinking, and InternVL2-8B, specifically focusing on their zero-shot performance on the MVP task. 
To ensure statistical significance, the evaluation was conducted across 2,000 diverse samples, utilizing a controlled masking configuration where the number of masked frames followed a 2:3:4 distribution at a fixed 1:2:1 ratio.
\begin{table}[ht]
\centering

\begin{tabularx}{\linewidth}{l >{\centering\arraybackslash}X >{\centering\arraybackslash}X}
\hline
\textbf{Model} & \textbf{Avg. Accuracy (\%)} & \textbf{Avg. Format Rate} \\ \hline
Qwen2.5-VL-7B  & 22.37                       & 1.0000                    \\
Qwen3-VL-8B    & 21.54                       & 0.7405                    \\
InternVL3.5-8B & 5.00                        & 0.3110                    \\ \hline
\end{tabularx}
\caption{\textbf{Performance Comparison of Different Base Models on MVP.}}
\label{tab:mvp_results}
\end{table}
Our empirical results reveal that the MVP task poses a substantial challenge for current state-of-the-art models, as evidenced by the relatively low average accuracy across the board. 
The complexity of temporal reasoning and visual reconstruction inherent in this task suggests that setting an overly aggressive masking strategy could impede the convergence of the model.
Consequently, these findings provide a critical heuristic for our training phase: to maintain a stable learning signal and avoid catastrophic forgetting or optimization difficulties, the mask length should be carefully calibrated and kept within a moderate range rather than being set to an excessively high value.

\section{Case Study}
\begin{figure*}[t]
\centering
\includegraphics[width=1.0\linewidth]{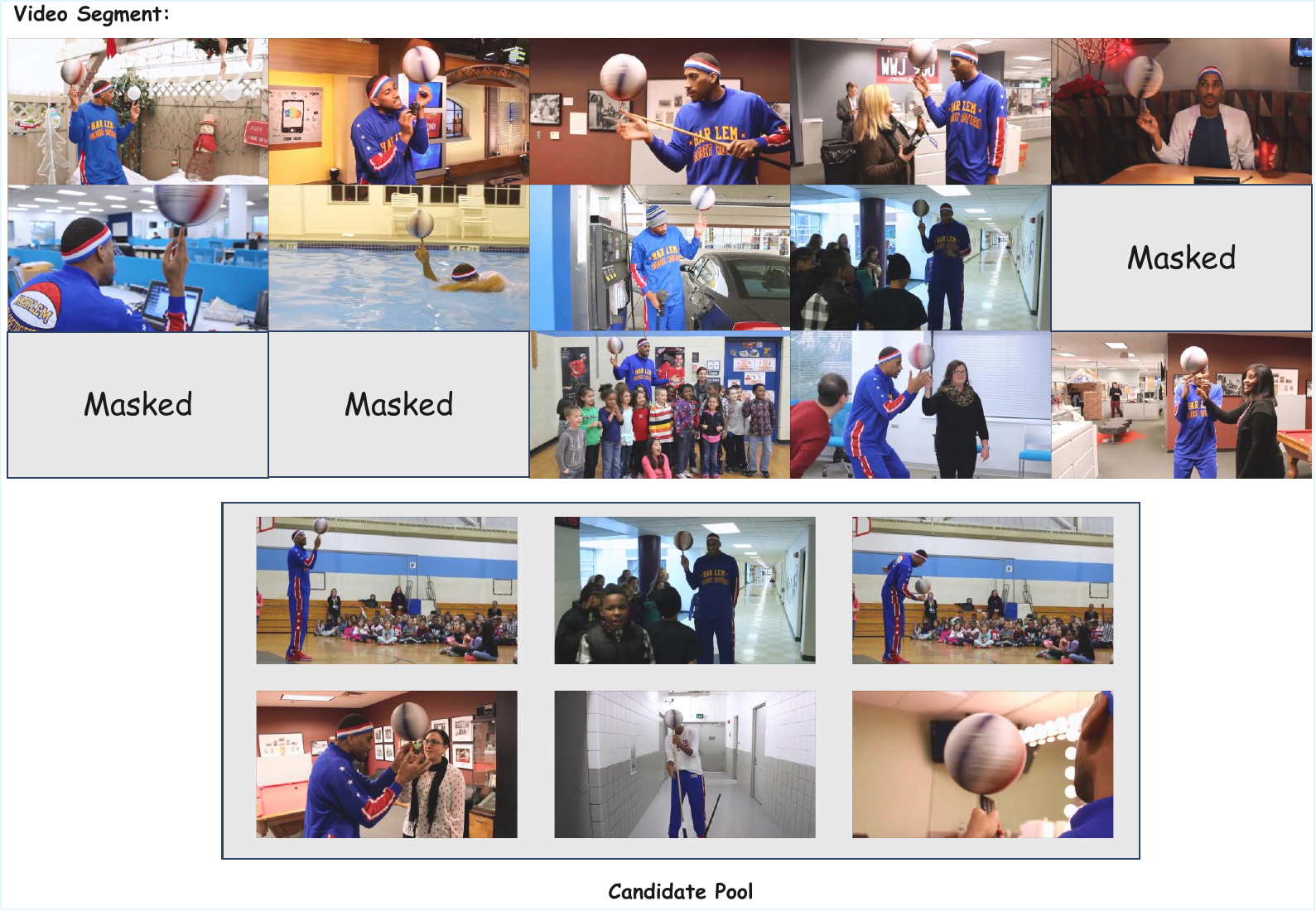}
\vspace{-10pt}
\caption{\textbf{Example of an MVP sample.} The answer is [b,a,c].}
\label{fig:example_mvp}
\vspace{-10pt}
\end{figure*}
Further visualization and comparison with base models are provided in this section.
We can see from most of the cases that after MVP training, the model focuses more on visual information, and reduces hallucinations that occur during the reasoning process.

\begin{figure*}[t]
\centering
\includegraphics[width=1.0\linewidth]{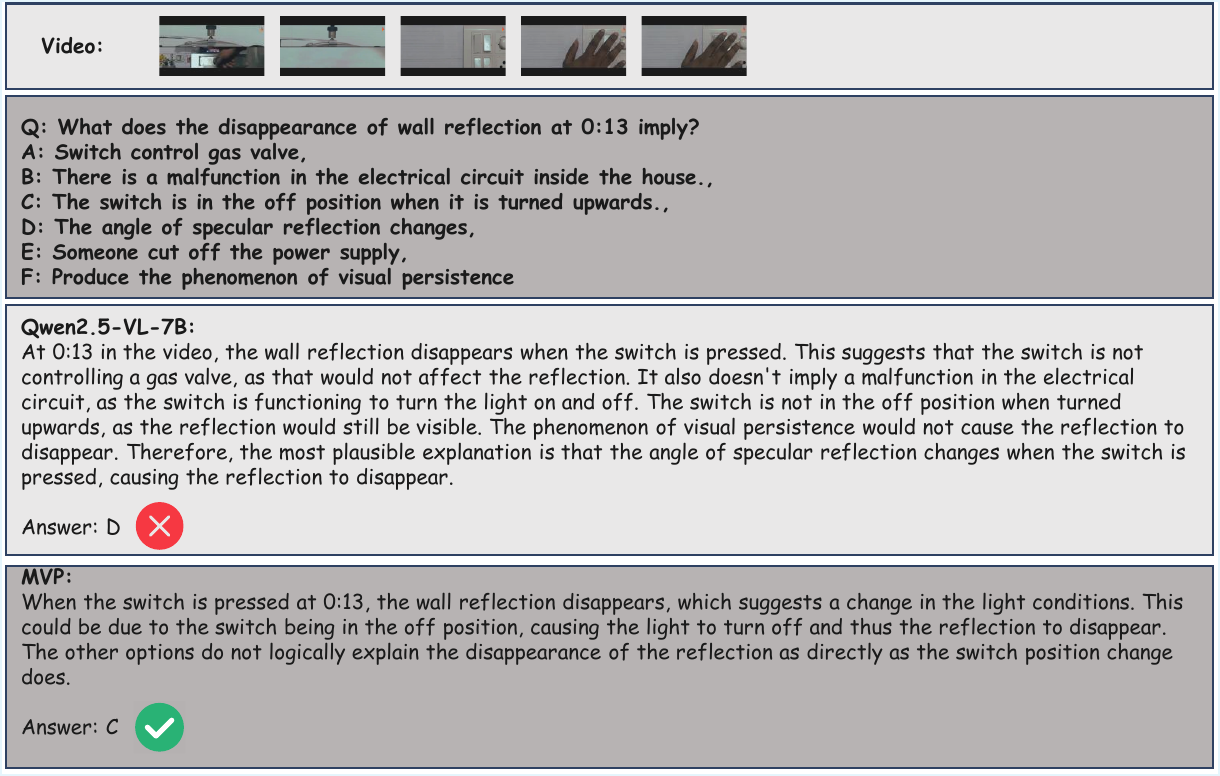}
\vspace{-10pt}
\caption{\textbf{An example of Qwen2.5-VL-Instruct before and after MVP training.}}
\label{fig:expl_qwen1}
\vspace{-10pt}
\end{figure*}

\begin{figure*}[t]
\centering
\includegraphics[width=1.0\linewidth]{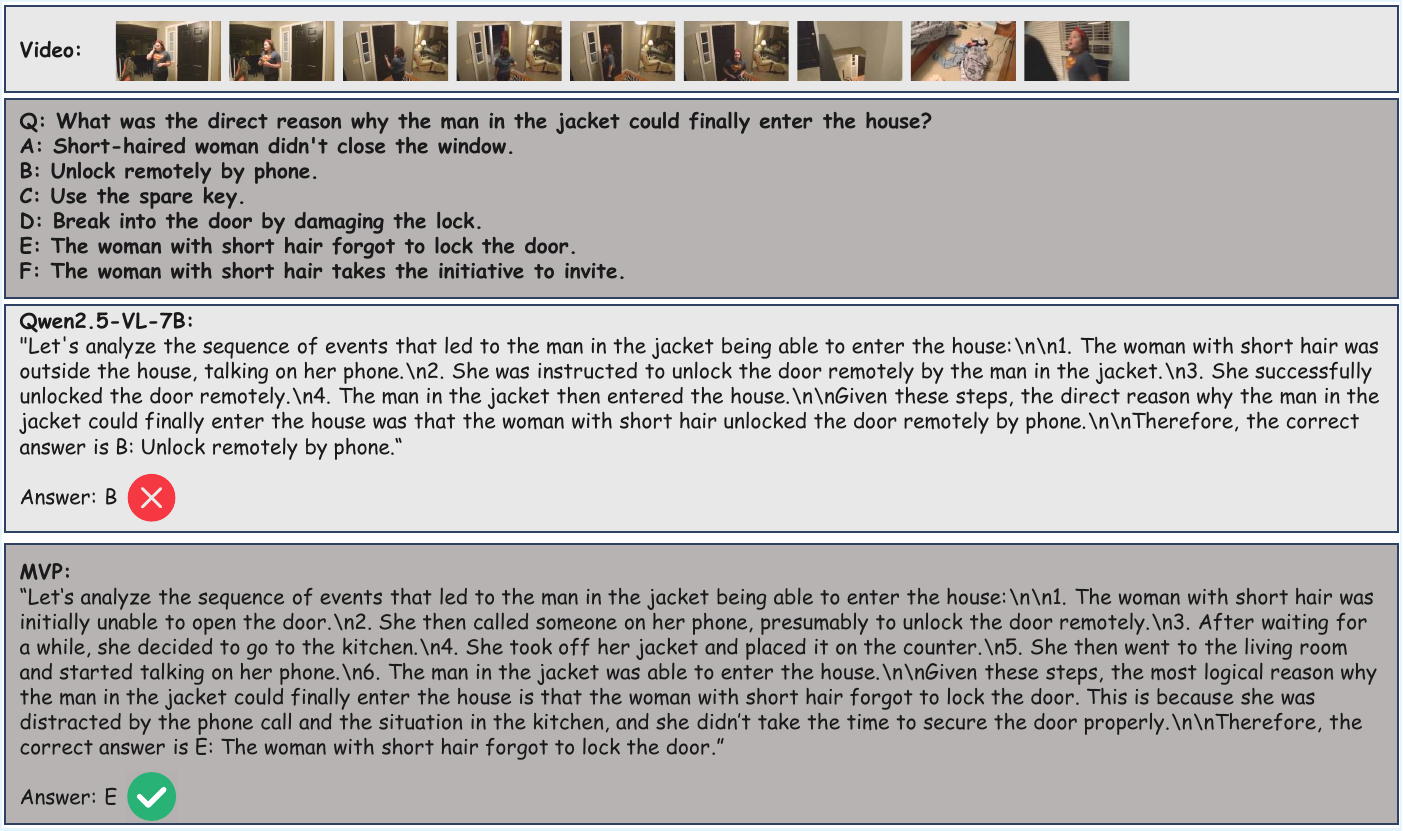}
\vspace{-10pt}
\caption{\textbf{An example of Qwen2.5-VL-Instruct before and after MVP training.}}
\label{fig:expl_qwen2}
\vspace{-10pt}
\end{figure*}

\begin{figure*}[t]
\centering
\includegraphics[width=1.0\linewidth]{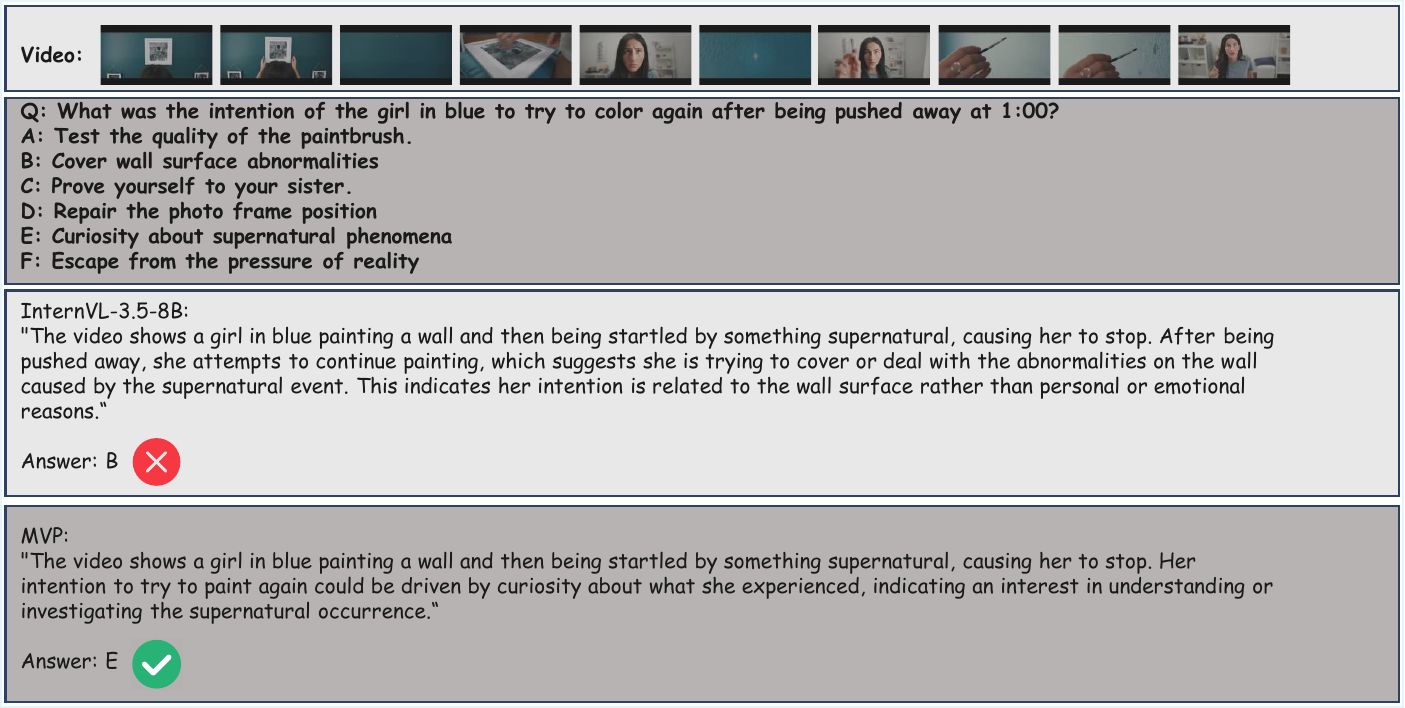}
\vspace{-10pt}
\caption{\textbf{An example of InternVL-3.5 before and after MVP training.}}
\label{fig:expl_intern}
\vspace{-10pt}
\end{figure*}

%% file: acl_latex.bbl
\begin{thebibliography}{39}
\providecommand{\natexlab}[1]{#1}

\bibitem[{An et~al.(2025)An, Xie, Yang, Zhang, Zhao, Cheng, Wang, Xu, Chen, Wu et~al.}]{an2025llava}
Xiang An, Yin Xie, Kaicheng Yang, Wenkang Zhang, Xiuwei Zhao, Zheng Cheng, Yirui Wang, Songcen Xu, Changrui Chen, Chunsheng Wu, and 1 others. 2025.
\newblock Llava-onevision-1.5: Fully open framework for democratized multimodal training.
\newblock \emph{arXiv preprint arXiv:2509.23661}.

\bibitem[{Bai et~al.(2025{\natexlab{a}})Bai, Cai, Chen, Chen, Chen, Cheng, Deng, Ding, Gao, Ge, Ge, Guo, Huang, Huang, Huang, Hui, Jiang, Li, Li, Li, Li, Lin, Lin, Liu, Liu, Liu, Liu, Liu, Liu, Lu, Luo, Lv, Men, Meng, Ren, Ren, Song, Sun, Tang, Tu, Wan, Wang, Wang, Wang, Wang, Xie, Xu, Xu, Xu, Yang, Yang, Yang, Yang, Yu, Zhang, Zhang, Zhang, Zheng, Zhong, Zhou, Zhou, Zhou, Zhu, and Zhu}]{bai2025qwen3vltechnicalreport}
Shuai Bai, Yuxuan Cai, Ruizhe Chen, Keqin Chen, Xionghui Chen, Zesen Cheng, Lianghao Deng, Wei Ding, Chang Gao, Chunjiang Ge, Wenbin Ge, Zhifang Guo, Qidong Huang, Jie Huang, Fei Huang, Binyuan Hui, Shutong Jiang, Zhaohai Li, Mingsheng Li, and 45 others. 2025{\natexlab{a}}.
\newblock \href {https://arxiv.org/abs/2511.21631} {Qwen3-vl technical report}.
\newblock \emph{Preprint}, arXiv:2511.21631.

\bibitem[{Bai et~al.(2025{\natexlab{b}})Bai, Chen, Liu, Wang, Ge, Song, Dang, Wang, Wang, Tang, Zhong, Zhu, Yang, Li, Wan, Wang, Ding, Fu, Xu, Ye, Zhang, Xie, Cheng, Zhang, Yang, Xu, and Lin}]{bai2025qwen25vltechnicalreport}
Shuai Bai, Keqin Chen, Xuejing Liu, Jialin Wang, Wenbin Ge, Sibo Song, Kai Dang, Peng Wang, Shijie Wang, Jun Tang, Humen Zhong, Yuanzhi Zhu, Mingkun Yang, Zhaohai Li, Jianqiang Wan, Pengfei Wang, Wei Ding, Zheren Fu, Yiheng Xu, and 8 others. 2025{\natexlab{b}}.
\newblock \href {https://arxiv.org/abs/2502.13923} {Qwen2.5-vl technical report}.
\newblock \emph{Preprint}, arXiv:2502.13923.

\bibitem[{Cheng et~al.(2025)Cheng, Ge, Wang, Ge, Liao, and Shan}]{cheng2025video}
Junhao Cheng, Yuying Ge, Teng Wang, Yixiao Ge, Jing Liao, and Ying Shan. 2025.
\newblock Video-holmes: Can mllm think like holmes for complex video reasoning?
\newblock \emph{arXiv preprint arXiv:2505.21374}.

\bibitem[{DeepSeek-AI et~al.(2025)DeepSeek-AI, Guo, Yang, Zhang, Song, Zhang, Xu, Zhu, Ma, Wang, Bi, Zhang, Yu, Wu, Wu, Gou, Shao, Li, Gao, Liu, Xue, Wang, Wu, Feng, Lu, Zhao, Deng, Zhang, Ruan, Dai, Chen, Ji, Li, Lin, Dai, Luo, Hao, Chen, Li, Zhang, Bao, Xu, Wang, Ding, Xin, Gao, Qu, Li, Guo, Li, Wang, Chen, Yuan, Qiu, Li, Cai, Ni, Liang, Chen, Dong, Hu, Gao, Guan, Huang, Yu, Wang, Zhang, Zhao, Wang, Zhang, Xu, Xia, Zhang, Zhang, Tang, Li, Wang, Li, Tian, Huang, Zhang, Wang, Chen, Du, Ge, Zhang, Pan, Wang, Chen, Jin, Chen, Lu, Zhou, Chen, Ye, Wang, Yu, Zhou, Pan, Li, Zhou, Wu, Ye, Yun, Pei, Sun, Wang, Zeng, Zhao, Liu, Liang, Gao, Yu, Zhang, Xiao, An, Liu, Wang, Chen, Nie, Cheng, Liu, Xie, Liu, Yang, Li, Su, Lin, Li, Jin, Shen, Chen, Sun, Wang, Song, Zhou, Wang, Shan, Li, Wang, Wei, Zhang, Xu, Li, Zhao, Sun, Wang, Yu, Zhang, Shi, Xiong, He, Piao, Wang, Tan, Ma, Liu, Guo, Ou, Wang, Gong, Zou, He, Xiong, Luo, You, Liu, Zhou, Zhu, Xu, Huang, Li, Zheng, Zhu, Ma, Tang, Zha, Yan, Ren, Ren, Sha, Fu, Xu, Xie, Zhang,
  Hao, Ma, Yan, Wu, Gu, Zhu, Liu, Li, Xie, Song, Pan, Huang, Xu, Zhang, and Zhang}]{deepseekai2025deepseekr1incentivizingreasoningcapability}
DeepSeek-AI, Daya Guo, Dejian Yang, Haowei Zhang, Junxiao Song, Ruoyu Zhang, Runxin Xu, Qihao Zhu, Shirong Ma, Peiyi Wang, Xiao Bi, Xiaokang Zhang, Xingkai Yu, Yu~Wu, Z.~F. Wu, Zhibin Gou, Zhihong Shao, Zhuoshu Li, Ziyi Gao, and 181 others. 2025.
\newblock \href {https://arxiv.org/abs/2501.12948} {Deepseek-r1: Incentivizing reasoning capability in llms via reinforcement learning}.
\newblock \emph{Preprint}, arXiv:2501.12948.

\bibitem[{Devlin et~al.(2019)Devlin, Chang, Lee, and Toutanova}]{devlin2019bert}
Jacob Devlin, Ming-Wei Chang, Kenton Lee, and Kristina Toutanova. 2019.
\newblock Bert: Pre-training of deep bidirectional transformers for language understanding.
\newblock In \emph{Proceedings of the 2019 conference of the North American chapter of the association for computational linguistics: human language technologies, volume 1 (long and short papers)}, pages 4171--4186.

\bibitem[{Feng et~al.(2025)Feng, Gong, Li, Guo, Wang, Peng, Wu, Zhang, Wang, and Yue}]{feng2025video}
Kaituo Feng, Kaixiong Gong, Bohao Li, Zonghao Guo, Yibing Wang, Tianshuo Peng, Junfei Wu, Xiaoying Zhang, Benyou Wang, and Xiangyu Yue. 2025.
\newblock Video-r1: Reinforcing video reasoning in mllms.
\newblock \emph{arXiv preprint arXiv:2503.21776}.

\bibitem[{Fu et~al.(2025{\natexlab{a}})Fu, Dai, Luo, Li, Ren, Zhang, Wang, Zhou, Shen, Zhang et~al.}]{fu2025video}
Chaoyou Fu, Yuhan Dai, Yongdong Luo, Lei Li, Shuhuai Ren, Renrui Zhang, Zihan Wang, Chenyu Zhou, Yunhang Shen, Mengdan Zhang, and 1 others. 2025{\natexlab{a}}.
\newblock Video-mme: The first-ever comprehensive evaluation benchmark of multi-modal llms in video analysis.
\newblock In \emph{Proceedings of the Computer Vision and Pattern Recognition Conference}, pages 24108--24118.

\bibitem[{Fu et~al.(2025{\natexlab{b}})Fu, Yang, Li, Wei, Xie, and Zheng}]{fu2025love}
Shenghao Fu, Qize Yang, Yuan-Ming Li, Xihan Wei, Xiaohua Xie, and Wei-Shi Zheng. 2025{\natexlab{b}}.
\newblock Love-r1: Advancing long video understanding with an adaptive zoom-in mechanism via multi-step reasoning.
\newblock \emph{arXiv preprint arXiv:2509.24786}.

\bibitem[{Gao et~al.(2017)Gao, Sun, Yang, and Nevatia}]{gao2017tall}
Jiyang Gao, Chen Sun, Zhenheng Yang, and Ram Nevatia. 2017.
\newblock Tall: Temporal activity localization via language query.
\newblock In \emph{Proceedings of the IEEE international conference on computer vision}, pages 5267--5275.

\bibitem[{Guo et~al.(2025)Guo, Wu, Zhu, Leng, Shi, Chen, Fan, Wang, Jiang, Wang, Chen, Huang, Lei, Yuan, Luo, Liu, Ye, Qian, Yan, Zhao, Peng, Li, Yuan, Wu, Cheng, Liu, Wang, Zeng, Liu, Qin, Ding, Xiao, Zhang, Zhang, Xiong, Peng, Chen, Li, Hu, Lin, Hu, Zhang, Wu, Li, Liu, Ling, Qin, Wang, He, Zhang, Yi, Liao, Huang, Zhang, Deng, Deng, Lin, Yuan, Li, Gou, Lou, Wei, Liu, Li, Zhu, Zhong, Li, Zhang, Wu, Li, Xiao, Lin, Yang, Wang, Ji, Hao, Shen, Li, Li, Wu, Zhu, Jiao, Feng, Chen, Duan, Liu, Zeng, Tang, Sun, Chen, Long, Feng, Zhan, Fang, Lu, Hua, Liu, Shen, Zhang, Shen, Wang, Pan, Zhang, Li, Li, Li, Shi, Han, Xiang, Chen, Chen, Li, Yan, Chi, Liu, Du, Wang, Pan, Chen, Chen, Wu, Yuan, Shuai, Tao, Zheng, Zhang, Zhang, Wang, Yang, Zhao, Xu, Liang, Yan, Zhong, Cao, Wu, Liu, Chang, Cai, Ao, Yang, Zhang, Zhong, Jia, Weng, Yu, Huang, Zhu, Yang, Wang, Long, Yin, Li, Zhu, Jia, Zhang, Liu, Zhang, Yang, Luo, Chen, Zhong, Xiao, Li, Wu, Wen, Du, Zhang, Ye, Wu, Liu, Yue, Zhou, Yuan, Xu, Yang, Zhang, Fang, Li, Ren, Xiong, Hong,
  Wang, Sun, Wang, Cai, Zha, An, Zhao, Xu, Chen, Wu, Zheng, Wang, Huang, Zhu, and Song}]{guo2025seed15vltechnicalreport}
Dong Guo, Faming Wu, Feida Zhu, Fuxing Leng, Guang Shi, Haobin Chen, Haoqi Fan, Jian Wang, Jianyu Jiang, Jiawei Wang, Jingji Chen, Jingjia Huang, Kang Lei, Liping Yuan, Lishu Luo, Pengfei Liu, Qinghao Ye, Rui Qian, Shen Yan, and 178 others. 2025.
\newblock \href {https://arxiv.org/abs/2505.07062} {Seed1.5-vl technical report}.
\newblock \emph{Preprint}, arXiv:2505.07062.

\bibitem[{He et~al.(2025)He, Qu, Li, Huang, Liu, and Cheng}]{he2025videossrvideoselfsupervisedreinforcement}
Zefeng He, Xiaoye Qu, Yafu Li, Siyuan Huang, Daizong Liu, and Yu~Cheng. 2025.
\newblock \href {https://arxiv.org/abs/2511.06281} {Videossr: Video self-supervised reinforcement learning}.
\newblock \emph{Preprint}, arXiv:2511.06281.

\bibitem[{Hu et~al.(2025{\natexlab{a}})Hu, Wu, Pu, Xiao, Zhang, Yue, Li, and Liu}]{hu2025videommmuevaluatingknowledgeacquisition}
Kairui Hu, Penghao Wu, Fanyi Pu, Wang Xiao, Yuanhan Zhang, Xiang Yue, Bo~Li, and Ziwei Liu. 2025{\natexlab{a}}.
\newblock \href {https://arxiv.org/abs/2501.13826} {Video-mmmu: Evaluating knowledge acquisition from multi-discipline professional videos}.
\newblock \emph{Preprint}, arXiv:2501.13826.

\bibitem[{Hu et~al.(2025{\natexlab{b}})Hu, Wu, Pu, Xiao, Zhang, Yue, Li, and Liu}]{hu2025video}
Kairui Hu, Penghao Wu, Fanyi Pu, Wang Xiao, Yuanhan Zhang, Xiang Yue, Bo~Li, and Ziwei Liu. 2025{\natexlab{b}}.
\newblock Video-mmmu: Evaluating knowledge acquisition from multi-discipline professional videos.
\newblock \emph{arXiv preprint arXiv:2501.13826}.

\bibitem[{Lei et~al.(2021)Lei, Berg, and Bansal}]{lei2021qvhighlightsdetectingmomentshighlights}
Jie Lei, Tamara~L. Berg, and Mohit Bansal. 2021.
\newblock \href {https://arxiv.org/abs/2107.09609} {Qvhighlights: Detecting moments and highlights in videos via natural language queries}.
\newblock \emph{Preprint}, arXiv:2107.09609.

\bibitem[{Liu et~al.(2025)Liu, Zhao, Xu, and Ghanem}]{liu2025bolt}
Shuming Liu, Chen Zhao, Tianqi Xu, and Bernard Ghanem. 2025.
\newblock Bolt: Boost large vision-language model without training for long-form video understanding.
\newblock In \emph{Proceedings of the Computer Vision and Pattern Recognition Conference}, pages 3318--3327.

\bibitem[{Liu et~al.(2024)Liu, Li, Liu, Wang, Ren, Li, Chen, Sun, and Hou}]{liu2024tempcompass}
Yuanxin Liu, Shicheng Li, Yi~Liu, Yuxiang Wang, Shuhuai Ren, Lei Li, Sishuo Chen, Xu~Sun, and Lu~Hou. 2024.
\newblock Tempcompass: Do video llms really understand videos?
\newblock \emph{arXiv preprint arXiv:2403.00476}.

\bibitem[{OpenAI et~al.(2024)OpenAI, :, Jaech, Kalai, Lerer, Richardson, El-Kishky, Low, Helyar, Madry, Beutel, Carney, Iftimie, Karpenko, Passos, Neitz, Prokofiev, Wei, Tam, Bennett, Kumar, Saraiva, Vallone, Duberstein, Kondrich, Mishchenko, Applebaum, Jiang, Nair, Zoph, Ghorbani, Rossen, Sokolowsky, Barak, McGrew, Minaiev, Hao, Baker, Houghton, McKinzie, Eastman, Lugaresi, Bassin, Hudson, Li, de~Bourcy, Voss, Shen, Zhang, Koch, Orsinger, Hesse, Fischer, Chan, Roberts, Kappler, Levy, Selsam, Dohan, Farhi, Mely, Robinson, Tsipras, Li, Oprica, Freeman, Zhang, Wong, Proehl, Cheung, Mitchell, Wallace, Ritter, Mays, Wang, Such, Raso, Leoni, Tsimpourlas, Song, von Lohmann, Sulit, Salmon, Parascandolo, Chabot, Zhao, Brockman, Leclerc, Salman, Bao, Sheng, Andrin, Bagherinezhad, Ren, Lightman, Chung, Kivlichan, O'Connell, Osband, Gilaberte, Akkaya, Kostrikov, Sutskever, Kofman, Pachocki, Lennon, Wei, Harb, Twore, Feng, Yu, Weng, Tang, Yu, Candela, Palermo, Parish, Heidecke, Hallman, Rizzo, Gordon, Uesato, Ward,
  Huizinga, Wang, Chen, Xiao, Singhal, Nguyen, Cobbe, Shi, Wood, Rimbach, Gu-Lemberg, Liu, Lu, Stone, Yu, Ahmad, Yang, Liu, Maksin, Ho, Fedus, Weng, Li, McCallum, Held, Kuhn, Kondraciuk, Kaiser, Metz, Boyd, Trebacz, Joglekar, Chen, Tintor, Meyer, Jones, Kaufer, Schwarzer, Shah, Yatbaz, Guan, Xu, Yan, Glaese, Chen, Lampe, Malek, Wang, Fradin, McClay, Pavlov, Wang, Wang, Murati, Bavarian, Rohaninejad, McAleese, Chowdhury, Chowdhury, Ryder, Tezak, Brown, Nachum, Boiko, Murk, Watkins, Chao, Ashbourne, Izmailov, Zhokhov, Dias, Arora, Lin, Lopes, Gaon, Miyara, Leike, Hwang, Garg, Brown, James, Shu, Cheu, Greene, Jain, Altman, Toizer, Toyer, Miserendino, Agarwal, Hernandez, Baker, McKinney, Yan, Zhao, Hu, Santurkar, Chaudhuri, Zhang, Fu, Papay, Lin, Balaji, Sanjeev, Sidor, Broda, Clark, Wang, Gordon, Sanders, Patwardhan, Sottiaux, Degry, Dimson, Zheng, Garipov, Stasi, Bansal, Creech, Peterson, Eloundou, Qi, Kosaraju, Monaco, Pong, Fomenko, Zheng, Zhou, McCabe, Zaremba, Dubois, Lu, Chen, Cha, Bai, He, Zhang, Wang,
  Shao, and Li}]{openai2024openaio1card}
OpenAI, :, Aaron Jaech, Adam Kalai, Adam Lerer, Adam Richardson, Ahmed El-Kishky, Aiden Low, Alec Helyar, Aleksander Madry, Alex Beutel, Alex Carney, Alex Iftimie, Alex Karpenko, Alex~Tachard Passos, Alexander Neitz, Alexander Prokofiev, Alexander Wei, Allison Tam, and 244 others. 2024.
\newblock \href {https://arxiv.org/abs/2412.16720} {Openai o1 system card}.
\newblock \emph{Preprint}, arXiv:2412.16720.

\bibitem[{Park et~al.(2025)Park, Na, Kim, and Kim}]{park2025deepvideo}
Jinyoung Park, Jeehye Na, Jinyoung Kim, and Hyunwoo~J Kim. 2025.
\newblock Deepvideo-r1: Video reinforcement fine-tuning via difficulty-aware regressive grpo.
\newblock \emph{arXiv preprint arXiv:2506.07464}.

\bibitem[{Radford et~al.(2021)Radford, Kim, Hallacy, Ramesh, Goh, Agarwal, Sastry, Askell, Mishkin, Clark et~al.}]{radford2021learning}
Alec Radford, Jong~Wook Kim, Chris Hallacy, Aditya Ramesh, Gabriel Goh, Sandhini Agarwal, Girish Sastry, Amanda Askell, Pamela Mishkin, Jack Clark, and 1 others. 2021.
\newblock Learning transferable visual models from natural language supervision.
\newblock In \emph{International conference on machine learning}, pages 8748--8763. PmLR.

\bibitem[{Rafailov et~al.(2023)Rafailov, Sharma, Mitchell, Manning, Ermon, and Finn}]{rafailov2023direct}
Rafael Rafailov, Archit Sharma, Eric Mitchell, Christopher~D Manning, Stefano Ermon, and Chelsea Finn. 2023.
\newblock Direct preference optimization: Your language model is secretly a reward model.
\newblock \emph{Advances in neural information processing systems}, 36:53728--53741.

\bibitem[{Shao et~al.(2024)Shao, Wang, Zhu, Xu, Song, Bi, Zhang, Zhang, Li, Wu et~al.}]{shao2024deepseekmath}
Zhihong Shao, Peiyi Wang, Qihao Zhu, Runxin Xu, Junxiao Song, Xiao Bi, Haowei Zhang, Mingchuan Zhang, YK~Li, Y~Wu, and 1 others. 2024.
\newblock Deepseekmath: Pushing the limits of mathematical reasoning in open language models, 2024.
\newblock \emph{URL https://arxiv. org/abs/2402.03300}, 2(3):5.

\bibitem[{Tang et~al.(2025)Tang, Han, Sun, Zhou, Zhang, Wei, Yuan, Zhang, Xu, and Sun}]{tang2025tspo}
Canhui Tang, Zifan Han, Hongbo Sun, Sanping Zhou, Xuchong Zhang, Xin Wei, Ye~Yuan, Huayu Zhang, Jinglin Xu, and Hao Sun. 2025.
\newblock Tspo: Temporal sampling policy optimization for long-form video language understanding.
\newblock \emph{arXiv preprint arXiv:2508.04369}.

\bibitem[{Tao et~al.(2025)Tao, Wang, Hua, Cao, and Xu}]{tao2025digdifferentialgroundingenhancing}
Zhou Tao, Shida Wang, Yongxiang Hua, Haoyu Cao, and Linli Xu. 2025.
\newblock \href {https://arxiv.org/abs/2512.12633} {Dig: Differential grounding for enhancing fine-grained perception in multimodal large language model}.
\newblock \emph{Preprint}, arXiv:2512.12633.

\bibitem[{Team et~al.(2025)Team, Hong, Yu, Gu, Wang, Gan, Tang, Cheng, Qi, Ji, Pan, Duan, Wang, Wang, Cheng, He, Su, Yang, Pan, Zeng, Wang, Chen, Shi, Pang, Zhang, Yin, Yang, Chen, Xu, Zhu, Chen, Chen, Chen, Lin, Wang, Chen, Lei, Gong, Pan, Liu, Xu, Zhang, Zheng, Yang, Zhong, Huang, Zhao, Xue, Tu, Meng, Zhang, Luo, Hao, Tong, Li, Jia, Liu, Zhang, Lyu, Fan, Huang, Wang, Xue, Wang, Wang, An, Du, Shi, Huang, Niu, Wang, Yue, Li, Zhang, Wang, Wang, Zhang, Xue, Hou, Du, Wang, Zhang, Liu, Xu, Li, Huang, Dong, and Tang}]{vteam2025glm45vglm41vthinkingversatilemultimodal}
V~Team, Wenyi Hong, Wenmeng Yu, Xiaotao Gu, Guo Wang, Guobing Gan, Haomiao Tang, Jiale Cheng, Ji~Qi, Junhui Ji, Lihang Pan, Shuaiqi Duan, Weihan Wang, Yan Wang, Yean Cheng, Zehai He, Zhe Su, Zhen Yang, Ziyang Pan, and 69 others. 2025.
\newblock \href {https://arxiv.org/abs/2507.01006} {Glm-4.5v and glm-4.1v-thinking: Towards versatile multimodal reasoning with scalable reinforcement learning}.
\newblock \emph{Preprint}, arXiv:2507.01006.

\bibitem[{Wang et~al.(2025{\natexlab{a}})Wang, Liu, Liu, Du, Kawaguchi, Wang, and Pang}]{wang2025fostering}
Haonan Wang, Hongfu Liu, Xiangyan Liu, Chao Du, Kenji Kawaguchi, Ye~Wang, and Tianyu Pang. 2025{\natexlab{a}}.
\newblock Fostering video reasoning via next-event prediction.
\newblock \emph{arXiv preprint arXiv:2505.22457}.

\bibitem[{Wang et~al.(2025{\natexlab{b}})Wang, Su, Ren, Lin, and Chen}]{wang2025pixelreasonerincentivizingpixelspace}
Haozhe Wang, Alex Su, Weiming Ren, Fangzhen Lin, and Wenhu Chen. 2025{\natexlab{b}}.
\newblock \href {https://arxiv.org/abs/2505.15966} {Pixel reasoner: Incentivizing pixel-space reasoning with curiosity-driven reinforcement learning}.
\newblock \emph{Preprint}, arXiv:2505.15966.

\bibitem[{Wang et~al.(2025{\natexlab{c}})Wang, He, Hong, Cheng, Zhang, Qi, Ding, Gu, Huang, Xu et~al.}]{wang2025lvbench}
Weihan Wang, Zehai He, Wenyi Hong, Yean Cheng, Xiaohan Zhang, Ji~Qi, Ming Ding, Xiaotao Gu, Shiyu Huang, Bin Xu, and 1 others. 2025{\natexlab{c}}.
\newblock Lvbench: An extreme long video understanding benchmark.
\newblock In \emph{Proceedings of the IEEE/CVF International Conference on Computer Vision}, pages 22958--22967.

\bibitem[{Wang et~al.(2025{\natexlab{d}})Wang, Gao, Gu, Pu, Cui, Wei, Liu, Jing, Ye, Shao, Wang, Chen, Zhang, Yang, Wang, Wei, Yin, Li, Cui, Chen, Ding, Tian, Wu, Xie, Li, Yang, Duan, Wang, Hou, Hao, Zhang, Li, Zhao, Duan, Deng, Fu, He, Wang, He, Shi, He, Xiong, Lv, Wu, Shao, Zhang, Deng, Qi, Ge, Guo, Zhang, Zhang, Cao, Lin, Tang, Gao, Huang, Gu, Lyu, Tang, Wang, Lv, Ouyang, Wang, Dou, Zhu, Lu, Lin, Dai, Su, Zhou, Chen, Qiao, Wang, and Luo}]{wang2025internvl35advancingopensourcemultimodal}
Weiyun Wang, Zhangwei Gao, Lixin Gu, Hengjun Pu, Long Cui, Xingguang Wei, Zhaoyang Liu, Linglin Jing, Shenglong Ye, Jie Shao, Zhaokai Wang, Zhe Chen, Hongjie Zhang, Ganlin Yang, Haomin Wang, Qi~Wei, Jinhui Yin, Wenhao Li, Erfei Cui, and 56 others. 2025{\natexlab{d}}.
\newblock \href {https://arxiv.org/abs/2508.18265} {Internvl3.5: Advancing open-source multimodal models in versatility, reasoning, and efficiency}.
\newblock \emph{Preprint}, arXiv:2508.18265.

\bibitem[{Wu et~al.(2024)Wu, Li, Chen, and Li}]{wu2024longvideobench}
Haoning Wu, Dongxu Li, Bei Chen, and Junnan Li. 2024.
\newblock Longvideobench: A benchmark for long-context interleaved video-language understanding.
\newblock \emph{Advances in Neural Information Processing Systems}, 37:28828--28857.

\bibitem[{Wu et~al.(2025)Wu, Zhang, Diao, Li, Lu, and Liu}]{wu2025visual}
Penghao Wu, Yushan Zhang, Haiwen Diao, Bo~Li, Lewei Lu, and Ziwei Liu. 2025.
\newblock Visual jigsaw post-training improves mllms.
\newblock \emph{arXiv preprint arXiv:2509.25190}.

\bibitem[{Xing et~al.(2025)Xing, Dong, Zang, Cao, Liang, Huang, Wang, Wu, and Lin}]{xing2025caprl}
Long Xing, Xiaoyi Dong, Yuhang Zang, Yuhang Cao, Jianze Liang, Qidong Huang, Jiaqi Wang, Feng Wu, and Dahua Lin. 2025.
\newblock Caprl: Stimulating dense image caption capabilities via reinforcement learning.
\newblock \emph{arXiv preprint arXiv:2509.22647}.

\bibitem[{Xu et~al.(2025)Xu, Gao, Li, Lu, Gan, Lai, Cao, Kang, Yang, and Dehghan}]{xu2025slowfast}
Mingze Xu, Mingfei Gao, Shiyu Li, Jiasen Lu, Zhe Gan, Zhengfeng Lai, Meng Cao, Kai Kang, Yinfei Yang, and Afshin Dehghan. 2025.
\newblock Slowfast-llava-1.5: A family of token-efficient video large language models for long-form video understanding.
\newblock \emph{arXiv preprint arXiv:2503.18943}.

\bibitem[{Yan et~al.(2025)Yan, Li, He, Yue, Zeng, Wang, Qiao, Wang, and Wang}]{yan2025videochat}
Ziang Yan, Xinhao Li, Yinan He, Zhengrong Yue, Xiangyu Zeng, Yali Wang, Yu~Qiao, Limin Wang, and Yi~Wang. 2025.
\newblock Videochat-r1. 5: Visual test-time scaling to reinforce multimodal reasoning by iterative perception.
\newblock \emph{arXiv preprint arXiv:2509.21100}.

\bibitem[{Yang et~al.(2025)Yang, Wen, Ding, Liu, Chu, Song, Rao, Yi, Li, Zang et~al.}]{yang2025kwai}
Biao Yang, Bin Wen, Boyang Ding, Changyi Liu, Chenglong Chu, Chengru Song, Chongling Rao, Chuan Yi, Da~Li, Dunju Zang, and 1 others. 2025.
\newblock Kwai keye-vl 1.5 technical report.
\newblock \emph{arXiv preprint arXiv:2509.01563}.

\bibitem[{Zhang et~al.(2025{\natexlab{a}})Zhang, Wang, Huang, Li, Zhu, and Yin}]{zhang2025vcapsbench}
Shi-Xue Zhang, Hongfa Wang, Duojun Huang, Xin Li, Xiaobin Zhu, and Xu-Cheng Yin. 2025{\natexlab{a}}.
\newblock Vcapsbench: A large-scale fine-grained benchmark for video caption quality evaluation.
\newblock \emph{arXiv preprint arXiv:2505.23484}.

\bibitem[{Zhang et~al.(2025{\natexlab{b}})Zhang, Wu, Li, Li, Ma, Liu, and Li}]{zhang2025llavavideovideoinstructiontuning}
Yuanhan Zhang, Jinming Wu, Wei Li, Bo~Li, Zejun Ma, Ziwei Liu, and Chunyuan Li. 2025{\natexlab{b}}.
\newblock \href {https://arxiv.org/abs/2410.02713} {Llava-video: Video instruction tuning with synthetic data}.
\newblock \emph{Preprint}, arXiv:2410.02713.

\bibitem[{Zhao et~al.(2025)Zhao, Zhang, Xie, Hu, Gan, Long, Hu, Chen, Li, Xu et~al.}]{zhao2025mmvu}
Yilun Zhao, Haowei Zhang, Lujing Xie, Tongyan Hu, Guo Gan, Yitao Long, Zhiyuan Hu, Weiyuan Chen, Chuhan Li, Zhijian Xu, and 1 others. 2025.
\newblock Mmvu: Measuring expert-level multi-discipline video understanding.
\newblock In \emph{Proceedings of the Computer Vision and Pattern Recognition Conference}, pages 8475--8489.

\bibitem[{Zhou et~al.(2024)Zhou, Shu, Zhao, Wu, Xiao, Yang, Xiong, Zhang, Huang, and Liu}]{zhou2024mlvu}
Junjie Zhou, Yan Shu, Bo~Zhao, Boya Wu, Shitao Xiao, Xi~Yang, Yongping Xiong, Bo~Zhang, Tiejun Huang, and Zheng Liu. 2024.
\newblock Mlvu: A comprehensive benchmark for multi-task long video understanding.
\newblock \emph{arXiv e-prints}, pages arXiv--2406.

\end{thebibliography}
